\documentclass{llncs}
\pdfoutput=1
\usepackage{graphicx}
\usepackage[table]{xcolor}
\usepackage{tikz}
\usepackage{comment}
\usepackage{amsmath,amssymb} 
\usepackage{color}
\usepackage[accsupp]{axessibility}  
\usepackage[width=122mm,left=12mm,paperwidth=146mm,height=193mm,top=12mm,paperheight=217mm]{geometry}
\usepackage[caption=false]{subfig}
\usepackage{xspace}
\usepackage{booktabs}
\usepackage{enumitem}
\usepackage{longtable}

\newcommand{\OM}{\text{DMS}\xspace}

\newcommand{\matlabel}[1]{\emph{#1}\xspace}
\definecolor{matcolor0}{rgb}{0.0,0.0,0.0}
\definecolor{matcolor1}{rgb}{0.737,0.737,0.538}
\definecolor{matcolor2}{rgb}{0.0,0.737,0.0}
\definecolor{matcolor3}{rgb}{0.737,0.737,0.0}
\definecolor{matcolor4}{rgb}{0.0,0.0,0.737}
\definecolor{matcolor5}{rgb}{0.737,0.0,0.737}
\definecolor{matcolor6}{rgb}{0.0,0.737,0.737}
\definecolor{matcolor7}{rgb}{0.945,0.945,0.945}
\definecolor{matcolor8}{rgb}{0.0,0.538,0.538}
\definecolor{matcolor9}{rgb}{0.882,0.0,0.0}
\definecolor{matcolor10}{rgb}{0.538,0.737,0.0}
\definecolor{matcolor11}{rgb}{0.882,0.737,0.0}
\definecolor{matcolor12}{rgb}{0.538,0.0,0.737}
\definecolor{matcolor13}{rgb}{0.538,0.737,0.737}
\definecolor{matcolor14}{rgb}{0.0,0.882,0.538}
\definecolor{matcolor15}{rgb}{0.882,0.737,0.737}
\definecolor{matcolor16}{rgb}{0.0,0.538,0.0}
\definecolor{matcolor17}{rgb}{0.737,0.538,0.0}
\definecolor{matcolor18}{rgb}{0.538,0.882,0.737}
\definecolor{matcolor19}{rgb}{0.737,0.538,0.737}
\definecolor{matcolor20}{rgb}{0.0,0.538,0.737}
\definecolor{matcolor21}{rgb}{0.737,0.882,0.0}
\definecolor{matcolor22}{rgb}{0.0,0.882,0.737}
\definecolor{matcolor23}{rgb}{0.737,0.882,0.737}
\definecolor{matcolor24}{rgb}{0.538,0.538,0.0}
\definecolor{matcolor25}{rgb}{0.882,0.538,0.0}
\definecolor{matcolor26}{rgb}{0.538,0.882,0.0}
\definecolor{matcolor27}{rgb}{0.882,0.882,0.0}
\definecolor{matcolor28}{rgb}{0.538,0.538,0.737}
\definecolor{matcolor29}{rgb}{0.882,0.538,0.737}
\definecolor{matcolor30}{rgb}{0.0,0.882,0.0}
\definecolor{matcolor31}{rgb}{0.737,0.0,0.882}
\definecolor{matcolor32}{rgb}{0.0,0.0,0.538}
\definecolor{matcolor33}{rgb}{0.737,0.0,0.0}
\definecolor{matcolor34}{rgb}{0.0,0.737,0.538}
\definecolor{matcolor35}{rgb}{0.737,0.0,0.538}
\definecolor{matcolor36}{rgb}{0.0,0.0,0.882}
\definecolor{matcolor37}{rgb}{0.882,0.737,0.538}
\definecolor{matcolor38}{rgb}{0.0,0.737,0.882}
\definecolor{matcolor39}{rgb}{0.737,0.737,0.882}
\definecolor{matcolor40}{rgb}{0.538,0.0,0.538}
\definecolor{matcolor41}{rgb}{0.882,0.0,0.538}
\definecolor{matcolor42}{rgb}{0.538,0.737,0.538}
\definecolor{matcolor43}{rgb}{0.882,0.882,0.737}
\definecolor{matcolor44}{rgb}{0.538,0.0,0.882}
\definecolor{matcolor45}{rgb}{0.882,0.0,0.882}
\definecolor{matcolor46}{rgb}{0.538,0.737,0.882}
\definecolor{matcolor47}{rgb}{0.882,0.737,0.882}
\definecolor{matcolor48}{rgb}{0.0,0.538,0.882}
\definecolor{matcolor49}{rgb}{0.737,0.538,0.538}
\definecolor{matcolor50}{rgb}{0.737,0.737,0.737}
\definecolor{matcolor51}{rgb}{0.737,0.882,0.538}
\definecolor{matcolor52}{rgb}{0.538,0.0,0.0}
\definecolor{matcolor53}{rgb}{0.737,0.538,0.882}
\definecolor{matcolor54}{rgb}{0.0,0.882,0.882}
\definecolor{matcolor55}{rgb}{0.737,0.882,0.882}
\definecolor{matcolor56}{rgb}{0.538,0.538,0.538}

\newcommand\crule[3][black]{\textcolor{#1}{\rule{#2}{#3}}}

\def\eg{\emph{e.g.}} 
\def\ie{\emph{i.e.}} 
 
\def\etal{\emph{et al.}}

\makeatletter
\newcommand{\printfnsymbol}[1]{%
  \textsuperscript{\@fnsymbol{#1}}%
}
\makeatother
\begin{document}
\pagestyle{headings}
\title{A Dense Material Segmentation Dataset for Indoor and Outdoor Scene Parsing}
\author{Paul Upchurch\thanks{These authors contributed equally to this work.} \and Ransen Niu\printfnsymbol{1}}
\institute{Apple Inc.}
\maketitle
\begin{abstract}

A key algorithm for understanding the world is material segmentation, which assigns a label (metal, glass, etc.) to each pixel. We find that a model trained on existing data underperforms in some settings and propose to address this with a large-scale dataset of 3.2 million dense segments on 44,560 indoor and outdoor images, which is 23x more segments than existing data. Our data covers a more diverse set of scenes, objects, viewpoints and materials, and contains a more fair distribution of skin types. We show that a model trained on our data outperforms a state-of-the-art model across datasets and viewpoints. We propose a large-scale scene parsing benchmark and baseline of 0.729 per-pixel accuracy, 0.585 mean class accuracy and 0.420 mean IoU across 46 materials.

\end{abstract}

\section{Introduction}

A goal of computer vision is to develop the cognitive ability to plan manipulation of something and predict how it will respond to stimuli. This is informed by the properties of what something is made of. Those properties can be discovered by segmenting a photograph into recognized materials. Material recognition can be understood through the science of material perception starting with Adelson's~\cite{thingsstuff} proposal to divide the world into \emph{things} (countable objects) and \emph{stuff} (materials). Adelson argued stuff is important because of its ubiquity in everyday life. Ritchie \etal~\cite{ritchie2021material} describe material perception in two parts. The first part is categorical recognition of what something is made of. The second part is recognizing material properties (\eg, glossy, flexible, sound absorbent, sticky) which tells us how something will feel or how it will interact with other objects. While Schwartz \etal~\cite{schwartz2019recognizing} proposed to recognize properties from local image patches we follow Bell \etal~\cite{minc} who segmented images by recognizing material classes.

Deep learning-based material recognition builds on some key developments. Sharan \etal~\cite{sharan2013recognizing} showed that people can recognize 10 kinds of materials in the wild~\cite{fmd} with 85\% accuracy. Bell \etal~\cite{opensurfaces}, following~\cite{labelme}, built an efficient annotation tool to create a large-scale material database from crowds and Internet photos. Next, Bell \etal~\cite{minc} introduced large-scale training data and a deep learning approach leading to material segmentation as a building-block for haptics, material assignment, robotic navigation, acoustic simulation and context-aware mixed reality~\cite{gao2016deep,park2018photoshape,schissler2017acoustic,zhao2017fully,brandao2016material,chen2020context}. Xiao \etal~\cite{upernet} introduced a multi-task scene parsing model which endows a photograph with a rich prediction of scene type, objects, object parts, materials and textures.

Despite widespread adoption of material segmentation, a lack of large-scale data means evaluation rests on the only large-scale segmentation dataset, OpenSurfaces~\cite{opensurfaces}. We find there is room for improvement and propose the Dense Material Segmentation dataset (DMS) which has 3.2 million segments across 44k densely annotated images, and
show empirically that our data leads to models which further close the gap between computer vision and human perception.

\begin{figure}[t]
\includegraphics[width=\linewidth]{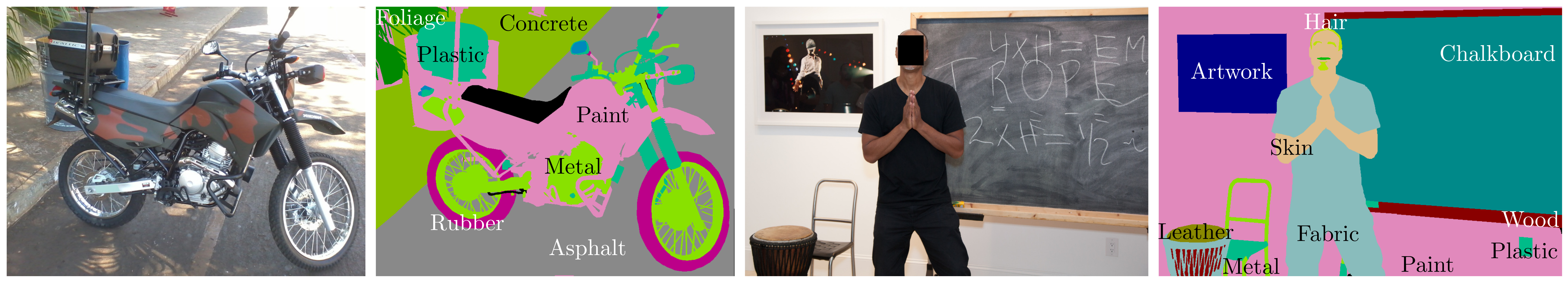}
\caption{{\bf Densely annotated materials.} Our annotations are full-scene, highly detailed and enable prediction of 46 kinds of materials.}
\label{fig:teaser}
\end{figure}

There are goals to consider for a material dataset.
First, we need a general-purpose set of material labels. We want to mimic human perception so we choose distinguishable materials even if they are of the same type. For example, we separate clear from opaque plastic rather than have a single label for all plastics. We define fine-grained labels which have useful properties, physical or otherwise. For example, a painted whiteboard surface has utility not found in a \matlabel{paint} label---it is appropriate for writing, cleaning and virtual content display. These functional properties come from how the material is applied rather than its physical structure. Ultimately we choose a set of 52 labels based on prior work and useful materials we found in photographs (details in Section~\ref{sec:dataset1}).

Following~\cite{schwartz2019recognizing}, we also want indoor and outdoor scenes. Counter-intuitively, this could be unnecessary. Material is recognizable regardless of where it occurs in the world, and deep learning methods aim to create a model which generalizes to unseen cases. Thus, an indoor residential dataset~\cite{opensurfaces} could be sufficient. We find this is not the case. In Section~\ref{sec:crossdataset} we show that a model trained on~\cite{opensurfaces} performs worse on outdoor scenes. This is a key finding which impacts all algorithms which use~\cite{opensurfaces} for training. We also show that a model trained on our dataset is consistent across indoor and outdoor scenes.

We want our database to support many scene parsing tasks so we need broad coverage of objects and scene attributes (which include activities, \eg, eating). In Section~\ref{sec:dataset2} we show that we achieve better coverage compared to~\cite{opensurfaces}.

We propose nine kinds of photographic types which distinguish different viewpoints and circumstances. Our motivation was to quantitatively evaluate cases where we had observed poor performance. This data can reveal new insights on how a model performs. We find that a state-of-the-art model underperforms in some settings whereas a model fit to our data performs well on all nine types.

Our final goal is to have diversity in skin types. Skin is associated with race and ethnicity so it is crucial to have fair representation across different types of skin. We compare our skin type data to OpenSurfaces~\cite{opensurfaces} in Section~\ref{sec:dataset2} and show our data has practical benefits for training in Section~\ref{sec:skin}.

The paper is organized as follows. In Section~\ref{sec:related} we review datasets. In Section~\ref{sec:dataset} we describe how we collected data to achieve our goals. In Section~\ref{sec:experiments} we compare our dataset to state-of-the-art data and a state-of-the-art model, study the impact of skin types on training, propose a material segmentation benchmark, and demonstrate material segmentation on real world photos.

In summary, our contributions are:
\begin{itemize}[topsep=3pt,itemsep=0pt]
\item We introduce \OM, a large-scale densely-annotated material segmentation dataset and show it is diverse with extensive analysis (Section~\ref{sec:dataset}).
\item We advance fairness toward skin types in material datasets (Section~\ref{sec:dataset2}).
\item We introduce photographic types which reveal new insights on prior work and show that a model fit to our data performs better across datasets and viewpoints compared to the state-of-the-art (Section~\ref{sec:crossdataset}).
\item We propose a new large-scale indoor and outdoor material segmentation benchmark of 46 materials and present a baseline result (Section~\ref{sec:baseline}).
\end{itemize}

\section{Related Work}
\label{sec:related}

{\bf Material Segmentation Datasets.} The largest dataset is OpenSurfaces~\cite{opensurfaces} which collected richly annotated polygons of residential indoor surfaces on 19k images, including 37 kinds of materials.
The largest material recognition dataset is the Materials in Context Database~\cite{minc} which is 3M point annotations of 23 kinds of materials across 437k images. This data enables material segmentation by CNN and a dense CRF tuned on OpenSurfaces segments.
The Local Materials Database~\cite{schwartz2019recognizing} collected segmentations, with the goal of studying materials using only local patches, of 16 kinds of materials across 5,845 images sourced from existing datasets.
The Light-Field Material Dataset~\cite{wang20164d} is 1,200 4D indoor and outdoor images of 12 kinds of materials.
The Multi-Illumination dataset~\cite{murmann2019dataset} captured 1,016 indoor scenes under 25 lighting conditions and annotated the images with 35 kinds of materials.
Table~\ref{tab:datasets} lists the largest datasets.

\begin{table}[t]
\centering
\caption{{\bf Large-scale datasets.} We propose a dataset with 23x more segments, more classes and 2.3x more images as the largest segment-annotated dataset.}
\label{tab:datasets}
\begin{tabular}{@{}lrlcrrl@{}}\toprule
 Dataset                                        & & Annotation         & Classes  & Images  & & Scenes\\\midrule
 OpenSurfaces~\cite{opensurfaces}               & & 137k segments      & 37       & 19,447  & & Indoor residential\\
 Materials in Context~\cite{minc}               & & 3M points          & 23       & 436,749 & & Home interior \& exterior\\
 Local Materials~\cite{schwartz2019recognizing} & & 9.4k segments      & 16       & 5,845   & & Indoor \& outdoor\\
 \OM (Ours)                                     & & 3.2M segments      & 52       & 44,560  & & Indoor \& outdoor\\
\bottomrule
\end{tabular}
\end{table}

Materials have appeared in purpose-built datasets. The Ground Terrain in Outdoor Scenes (GTOS) database~\cite{gtos} and GTOS-mobile~\cite{gtos2} are 30k images of hundreds of instances of 40 kinds of ground materials and 81 videos of 31 kinds of ground materials, respectively. The Materials in Paintings dataset~\cite{van2021materials} is bounding box annotations and extracted segmentations on 19k paintings of 15 kinds of materials depicted by artists, partly distinguished into 50 fine-grained categories.
COCO-Stuff~\cite{cocostuff} is segmentations of 91 kinds of stuff on 164k COCO~\cite{coco} images. While this is a source of material data, it is not a general-purpose material dataset because important surfaces (\eg, objects labeled in COCO) are not assigned material labels.
ClearGrasp~\cite{cleargrasp} is a dataset of 50k synthetic and 286 real RGB-D images of glass objects built for robotic manipulation of transparent objects.
The Glass Detection Dataset~\cite{mei2020don} is 3,916 indoor and outdoor images of segmented glass surfaces.
The Mirror Segmentation Dataset~\cite{yang2019my} is 4,018 images with segmented mirror surfaces across indoor and outdoor scenes.
Fashionpedia~\cite{fashionpedia} is a database of segmented clothing images of which 10k are annotated with fashion attributes which include fine-grained clothing materials.
Figaro~\cite{figaro} is 840 images of people with segmented hair distinguished into 7 kinds of hairstyles.

{\bf Categorical Material Names.}
Bell \etal~\cite{opensurfaces} created a set of names by asking annotators to enter free-form labels which were merged into a list of material names.
This approach is based on the appearance of surfaces as perceived by the annotators. Schwartz \etal~\cite{schwartz2019recognizing} created a three-level hierarchy of material names where materials are organized by their physical properties. Some categories were added for materials which could not be placed in the hierarchy. In practice, both approaches resulted in a similar set of entry-level~\cite{ordonez2013large} names which also closely agree with prior studies of categorical materials in Internet images~\cite{fmd,hu2011toward}.

\section{Data Collection}
\label{sec:dataset}

\OM is a set of dense polygon annotations of 52 material classes across 44,560 images, which are a subset of OpenImages~\cite{openimages2}. We followed a four step process. First, a set of labels was defined. Next, a large set of images was studied by people and algorithms to select images for annotation. Next, the selected images were fully segmented and labeled by a human annotator. Finally, each segmented image was relabeled by multiple people and a final label map was created by fusing all labels. The last three steps were followed multiple times.

\subsection{Material Labels}
\label{sec:dataset1}

We choose to predefine a label set which is the approach of COCO-Stuff~\cite{cocostuff}. This encourages annotators to create consistent labels suitable for machine learning. We instructed annotators to assign \matlabel{not on list} to recognized materials which do not fit in any category and \matlabel{I cannot tell} to unknown and unrecognizable surfaces (\eg, watermarks and under-/over-saturated pixels).

We defined a label set based on appearance, which is the approach of OpenSurfaces~\cite{opensurfaces}. A label can represent a solid substance (\eg, wood), a distinctive arrangement of substances (\eg, brickwork), a liquid (\eg, water) or a useful non-material (\eg, sky).
We used 35 labels from OpenSurfaces and \matlabel{asphalt} from~\cite{schwartz2019recognizing}.

\begin{figure}[t]
\subfloat[Asphalt]{\includegraphics[height=7.5ex]{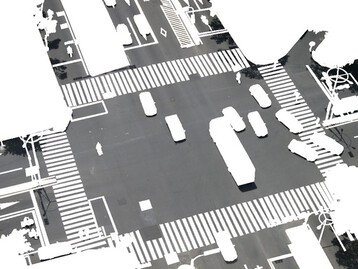}}\hfill
\subfloat[Bone]{\includegraphics[height=7.5ex]{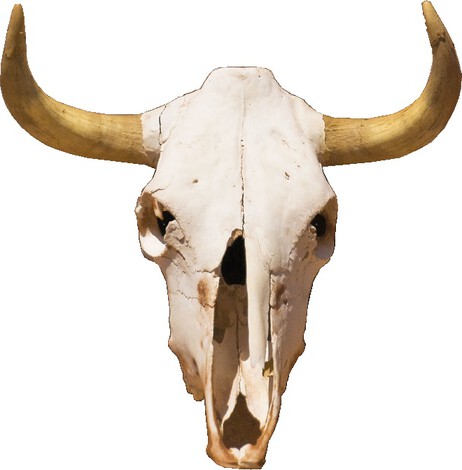}}\hfill
\subfloat[Brick]{\includegraphics[height=7.5ex]{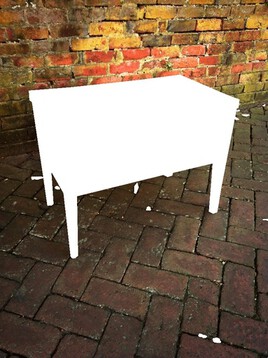}}\hfill
\subfloat[Eng. stone]{\includegraphics[height=7.5ex]{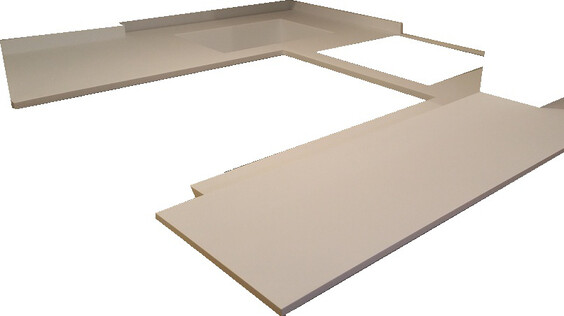}}\hfill
\subfloat[Fabric]{\includegraphics[height=7.5ex]{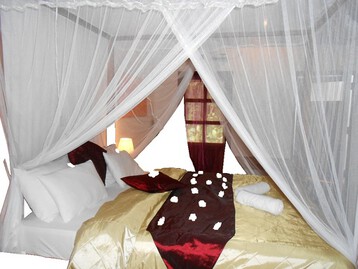}}\hfill
\subfloat[Carpet]{\includegraphics[height=7.5ex]{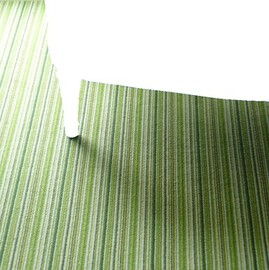}}\hfill
\subfloat[Ceiling tile]{\includegraphics[height=7.5ex]{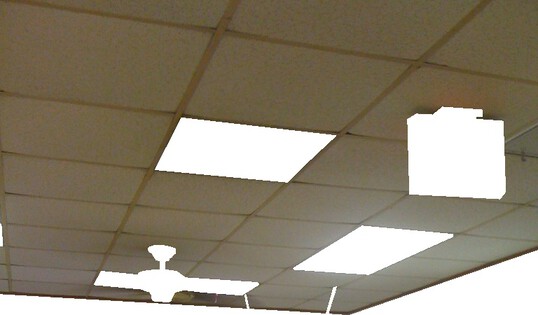}}\hfill
\subfloat[Ceramic]{\includegraphics[height=7.5ex]{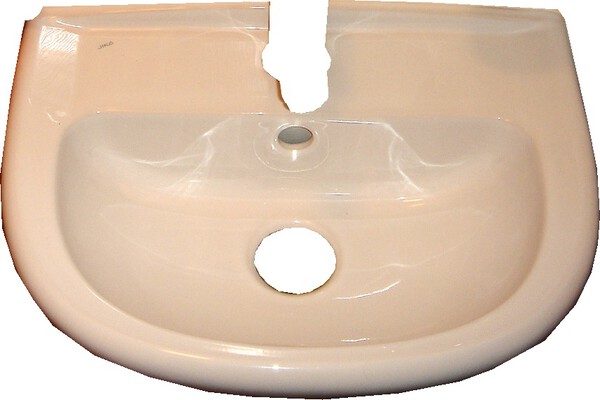}}\hfill
\subfloat[Wax]{\includegraphics[height=7.5ex]{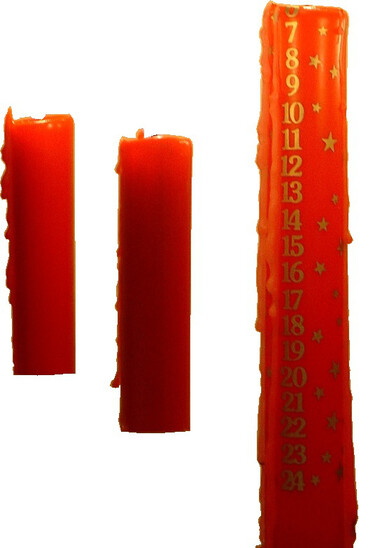}}

\subfloat[Wallpaper]{\includegraphics[height=7.5ex]{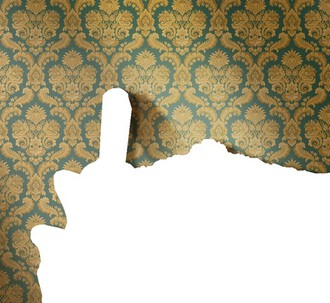}}\hfill
\subfloat[Clear plastic]{\includegraphics[height=7.5ex]{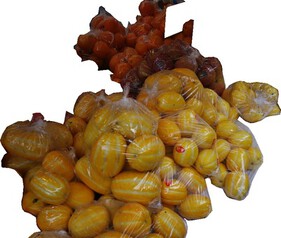}}\hfill
\subfloat[Plastic]{\includegraphics[height=7.5ex]{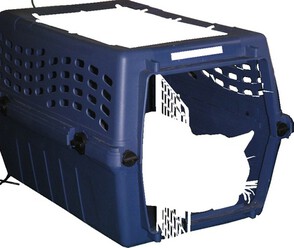}}\hfill
\subfloat[Concrete]{\includegraphics[height=7.5ex]{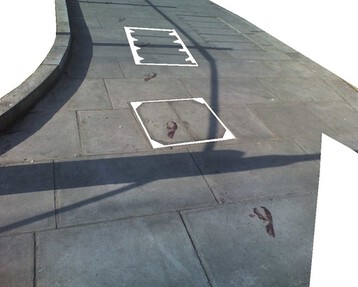}}\hfill
\subfloat[Artwork]{\includegraphics[height=7.5ex]{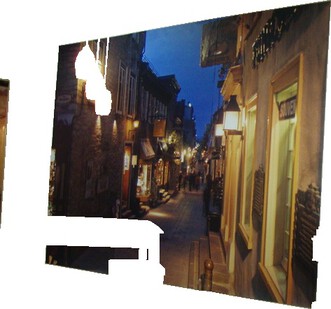}}\hfill
\subfloat[Cardboard]{\includegraphics[height=7.5ex]{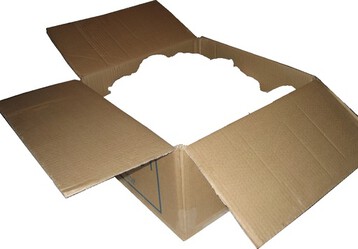}}\hfill
\subfloat[Chalkboard]{\includegraphics[height=7.5ex]{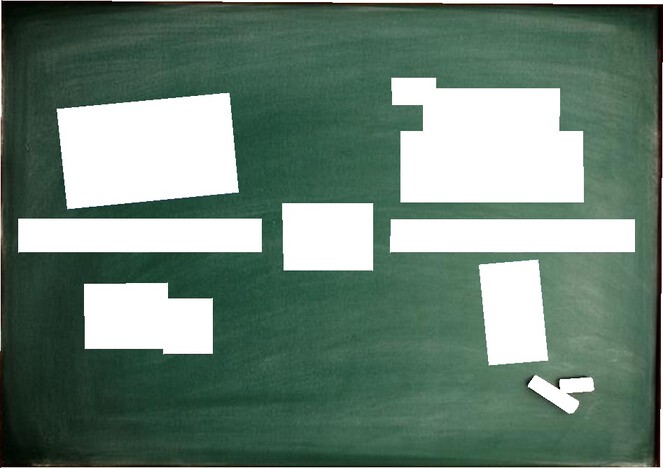}}\hfill
\subfloat[Fiberglass]{\includegraphics[height=7.5ex]{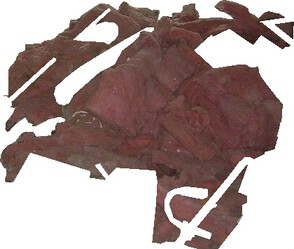}}\hfill
\subfloat[Rubber]{\includegraphics[height=7.5ex]{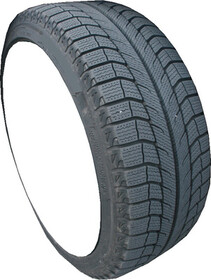}}

\subfloat[Fur]{\includegraphics[height=7.5ex]{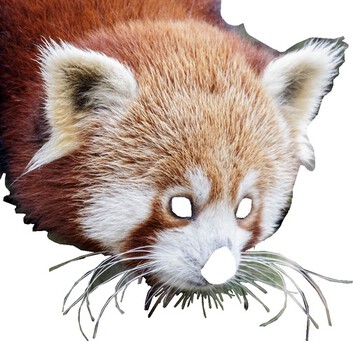}}\hfill
\subfloat[Foliage]{\includegraphics[height=7.5ex]{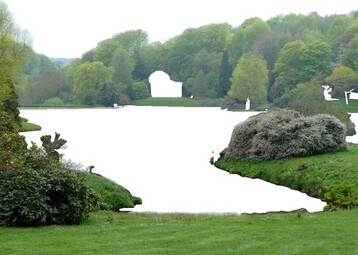}}\hfill
\subfloat[Food]{\includegraphics[height=7.5ex]{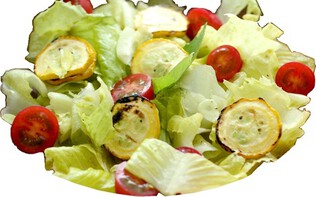}}\hfill
\subfloat[Hair]{\includegraphics[height=7.5ex]{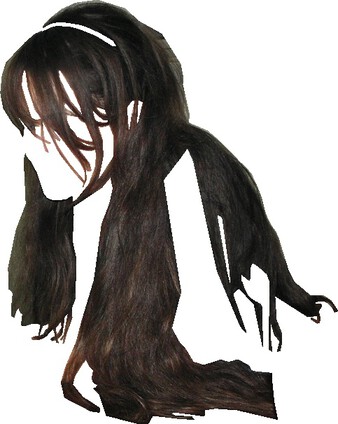}}\hfill
\subfloat[Cork]{\includegraphics[height=7.5ex]{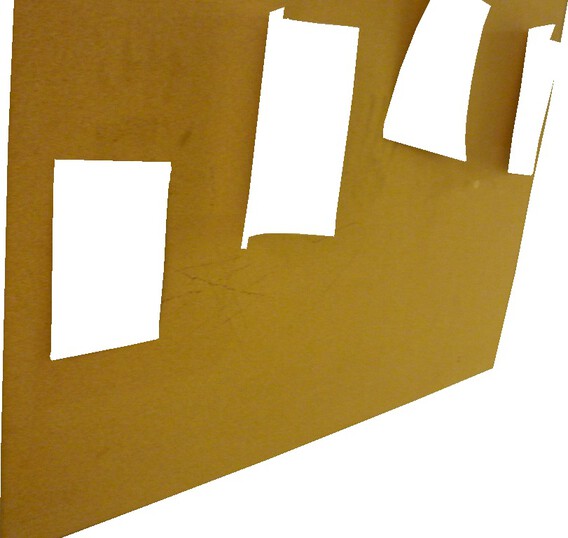}}\hfill
\subfloat[Fire]{\includegraphics[height=7.5ex]{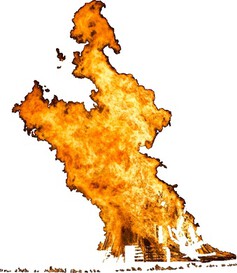}}\hfill
\subfloat[Gemstone]{\includegraphics[height=7.5ex]{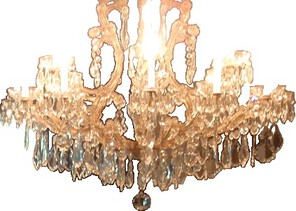}}\hfill
\subfloat[Glass]{\includegraphics[height=7.5ex]{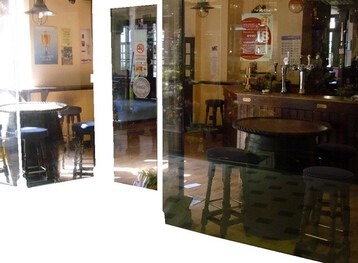}}\hfill
\subfloat[Ice]{\includegraphics[height=7.5ex]{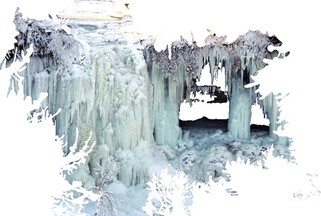}}

\subfloat[Paper]{\includegraphics[height=7.3ex]{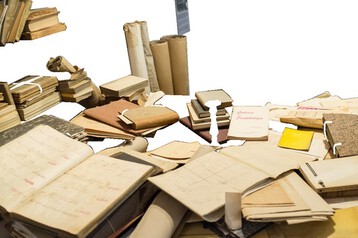}}\hfill
\subfloat[Leather]{\includegraphics[height=7.3ex]{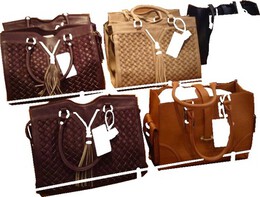}}\hfill
\subfloat[Liquid]{\includegraphics[height=7.3ex]{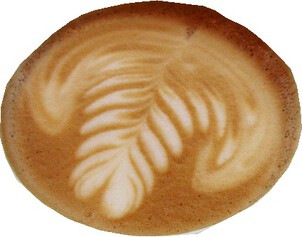}}\hfill
\subfloat[Metal]{\includegraphics[height=7.3ex]{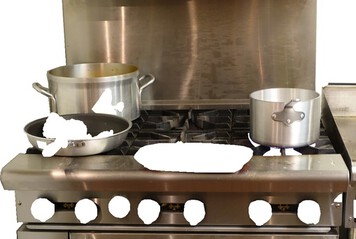}}\hfill
\subfloat[Mirror]{\includegraphics[height=7.3ex]{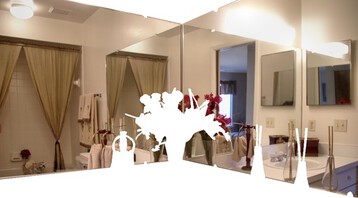}}\hfill
\subfloat[Paint]{\includegraphics[height=7.3ex]{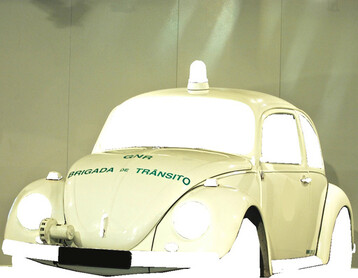}}\hfill
\subfloat[Pearl]{\includegraphics[height=7.3ex]{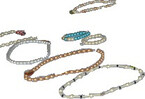}}\hfill
\subfloat[Sponge]{\includegraphics[height=7.3ex]{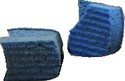}}

\subfloat[Soap]{\includegraphics[height=7.5ex]{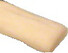}}\hfill
\subfloat[Clutter]{\includegraphics[height=7.5ex]{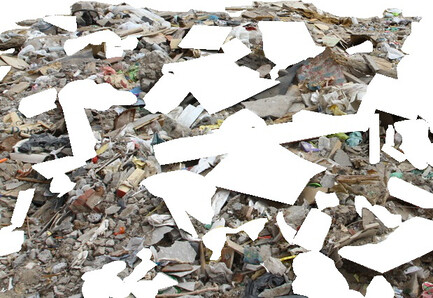}}\hfill
\subfloat[Wicker]{\includegraphics[height=7.5ex]{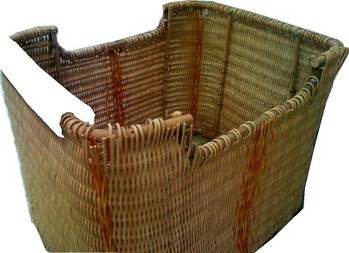}}\hfill
\subfloat[Snow]{\includegraphics[height=7.5ex]{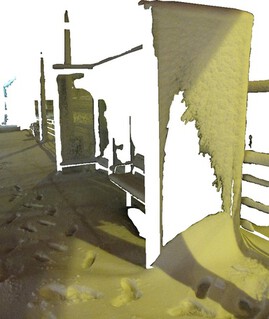}}\hfill
\subfloat[Sand]{\includegraphics[height=7.5ex]{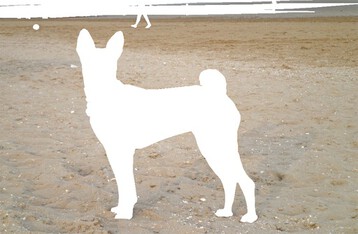}}\hfill
\subfloat[Skin]{\includegraphics[height=7.5ex]{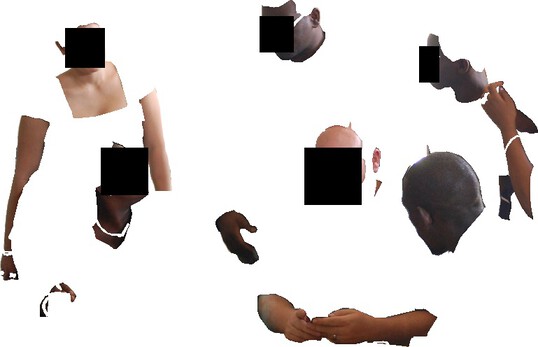}}\hfill
\subfloat[Sky]{\includegraphics[height=7.5ex]{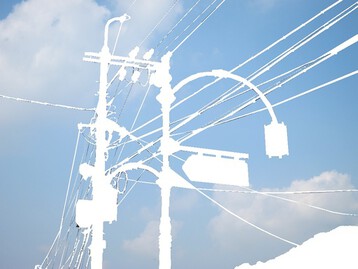}}\hfill
\subfloat[Soil]{\includegraphics[height=7.5ex]{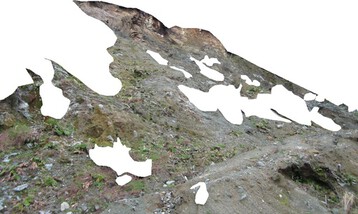}}

\subfloat[Nat. stone]{\includegraphics[height=7.5ex]{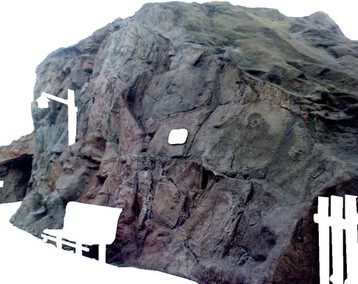}}\hfill
\subfloat[Pol. stone]{\includegraphics[height=7.5ex]{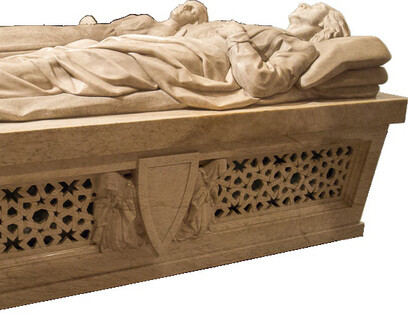}}\hfill
\subfloat[Styrofoam]{\includegraphics[height=7.5ex]{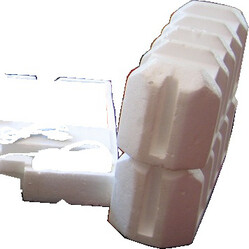}}\hfill
\subfloat[Tile]{\includegraphics[height=7.5ex]{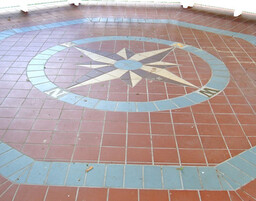}}\hfill
\subfloat[Water]{\includegraphics[height=7.5ex]{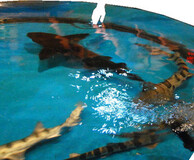}}\hfill
\subfloat[Whiteboard]{\includegraphics[height=7.5ex]{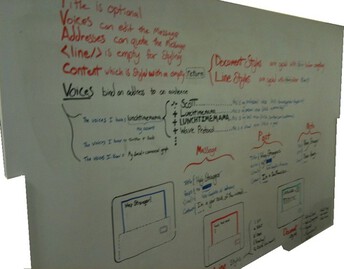}}\hfill
\subfloat[Wood]{\includegraphics[height=7.5ex]{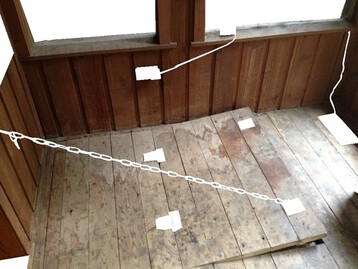}}\hfill
\subfloat[Tree wood]{\includegraphics[height=7.5ex]{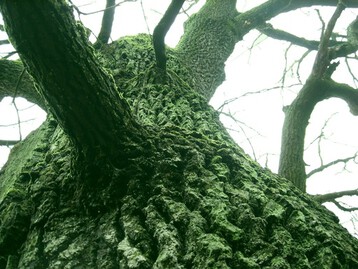}}\hfill
\subfloat[Animal skin]{\includegraphics[height=7.5ex]{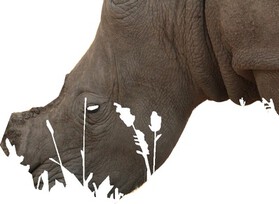}}
\caption{{\bf Material labels.} For each label we show a cut-out example.}
\label{fig:matlabels}
\end{figure}

We added 2 fine-grained people and animal categories (\matlabel{bone} and \matlabel{animal skin}). We introduced 3 labels for workplaces (\matlabel{ceiling tile}, \matlabel{whiteboard} and \matlabel{fiberglass wool}), 6 for indoor scenes (\matlabel{artwork}, \matlabel{clutter}, \matlabel{non-water liquid}, \matlabel{soap}, \matlabel{pearl} and \matlabel{gemstone}) and 4 for outdoors (\matlabel{sand}, \matlabel{snow}, \matlabel{ice} and \matlabel{tree wood}). \matlabel{Artwork} identifies an imitative surface which is photographic or fine art---affording further analysis by Materials In Paintings~\cite{van2021materials}. \matlabel{Clutter} is a region of visually indistinguishable manufactured stuff (typically a mixture of metal, plastic and paper) which occurs in trash piles. Lastly, we defined a label called \matlabel{engineered stone} for artificial surfaces which imitate stone, which includes untextured and laminated solid surfaces.
See Figure~\ref{fig:matlabels} for an example of each label.

\subsection{Image Selection}
\label{sec:dataset2}

\begin{figure}[t]
\centering
\includegraphics[width=\linewidth]{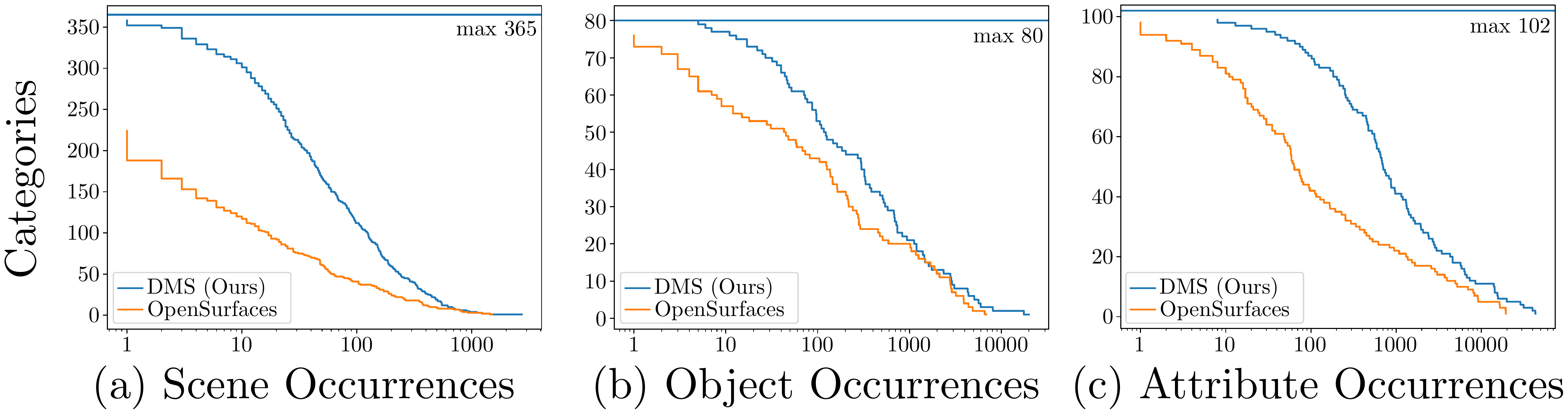}
\caption{{\bf Image diversity.} We plot number of categories ($y$-axis) vs. occurrence in images (log-scale $x$-axis) of Places365 scene type (a), COCO objects (b), and SUN attributes (c). Our dataset ({\it blue}) is larger, more diverse and more balanced across categories (higher slope) compared to the largest segmentation dataset ({\it orange}).}
\label{fig:occurrences}
\end{figure}

Bell \etal~\cite{minc} found that a balanced set of material labels can achieve nearly the same performance as a 9x larger imbalanced set. Since we collect dense annotations we cannot directly balance classes. Instead, we searched 191k images for rare materials and assumed common materials would co-occur. Furthermore, we ran Detectron~\cite{detectron} to detect COCO~\cite{coco} objects, and Places365~\cite{places365} to classify scenes and recognize SUN~\cite{sunattributes} attributes. EXIF information was used to infer country. These detections were used to select images of underrepresented scenes, objects and countries.
Figure~\ref{fig:occurrences} compares the diversity of the 45k images in \OM to the 19k images in OpenSurfaces by a plot of the number of categories, $y$, which have at least $x$ occurrences. Occurrences of scene type, object and SUN attribute are plotted. Note that the $x$-axis is logarithmic scale. We find our dataset is more diverse having more classes present in greater amounts (more than can be explained by the 2.24x difference in image count).

We balance the distribution of skin appearance in \OM so that algorithms trained with our data perform well on all kinds of skin~\cite{gendershades}. We use Fitzpatrick~\cite{fitzpatrick} skin type to categorize skin into 3 groups, inspired by an approach used by~\cite{towardsfairerdatasets}. We ran the DLIB~\cite{dlib} face detector and labeled a subset of the faces. Our 157 manual annotations were used to calibrate a preexisting face attribute predictor (trained on a different dataset) which was then used to predict skin types for the rest of \OM. We found that the ratio of the largest group to the smallest was 9.4. Next, we selected images which would increase the most underrepresented skin type group and found this reduced the ratio to 2.2. We calibrated the same detector for OpenSurfaces faces and measured its ratio as 10.4. According to the findings of~\cite{gendershades}, we expect skin classifiers trained on OpenSurfaces would underperform on dark skin. Table~\ref{tab:skintypes} shows the distribution of skin types.

We used Places365 scene type detection to select outdoor images but we found this did not lead to outdoor materials. We took two steps to address this. First, we annotated our images with one of nine \emph{photographic types} which distinguish outdoor from indoor from unreal images. Table~\ref{tab:photographic_type} shows the annotated types. Next, we used these labels to select outdoor scenes and underrepresented viewpoints. This was effective---growing the dataset by 17\% more than doubled 9 kinds of outdoor materials: \matlabel{ice} (3x), \matlabel{sand} (4.4x), \matlabel{sky} (8x), \matlabel{snow} (9.5x), \matlabel{soil} (3x), \matlabel{natural stone} (2.4x), \matlabel{water} (2.5x), \matlabel{tree wood} (2.3x) and \matlabel{asphalt} (9.2x).

\begin{table}[t]
\centering
\caption{{\bf Skin types.} We report estimated occurrences. Our dataset has 12x more occurrences of the smallest group and 4.8x more fair representation by ratio.}
\label{tab:skintypes}
\begin{tabular}{@{}lrr@{}}\toprule
                                     & OpenSurfaces  & \OM (Ours)\\\midrule
  Type I-II (light)                  & 2,332         & 4,535\\
  Type III-IV (medium)               & 3,889         & 9,776\\
  Type V-VI (dark)                   &   375         & 5,899\\\midrule
  Ratio of largest to smallest group & 10.37\,:\,1   & 2.16\,:\,1\\
\bottomrule
\end{tabular}
\end{table}

\begin{table}[t]
\centering
\caption{{\bf Photographic types.} Our data contains indoor views ({\it top}), outdoor views ({\it middle}), and close-up and unusual views ({\it bottom}).}
\label{tab:photographic_type}
\begin{minipage}[c]{0.69\linewidth}
\begin{tabular}{@{}lp{3mm}r@{}}\toprule
 Photographic Type                                           & & Images \\\midrule
 An area with visible enclosure                              & & 16,013 \\
 A collection of indoor things                               & &  6,064 \\
 A tightly cropped indoor thing                              & &  2,634 \\\midrule
 A ground-level view of reachable outdoor things             & &  3,127 \\
 A tightly cropped outdoor thing                             & &  1,196 \\
 Distant unreachable outdoor things                          & &    971 \\\midrule
 A real surface without context                              & &    847 \\
 Not a real photo                                            & &    805 \\
 An obstructed or distorted view                             & &    204 \\
\bottomrule
\end{tabular}
\end{minipage}
\begin{minipage}[c]{0.275\linewidth}
  \includegraphics[height=10.9ex]{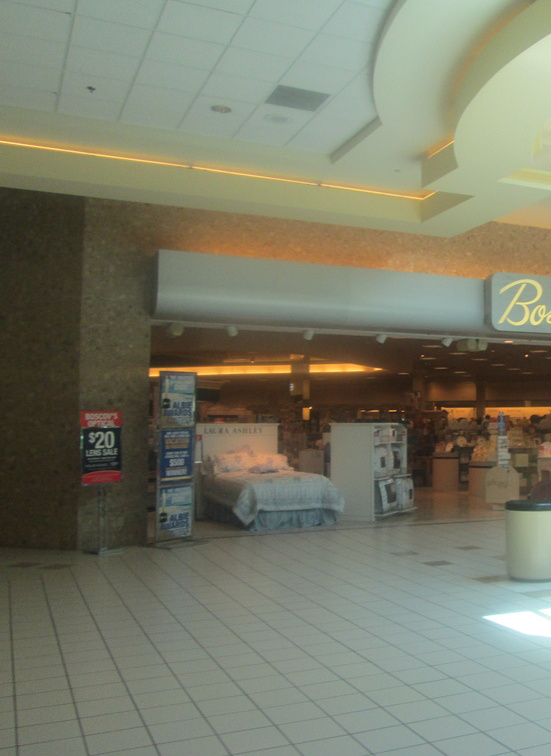}\hfill
  \includegraphics[height=10.9ex]{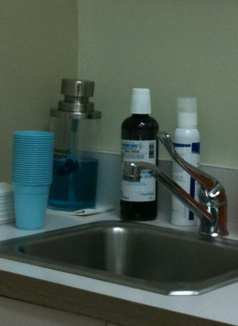}\hfill
  \includegraphics[height=10.9ex]{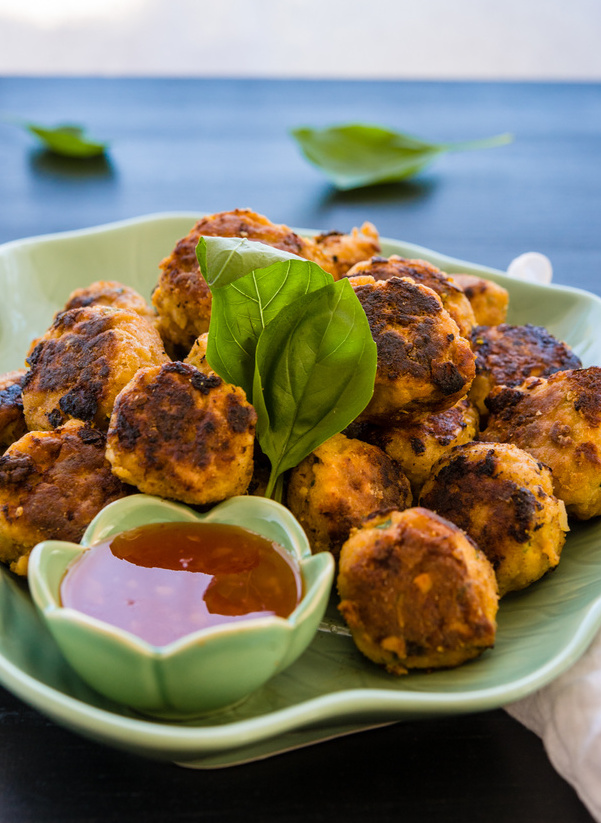}
  \includegraphics[height=10.9ex]{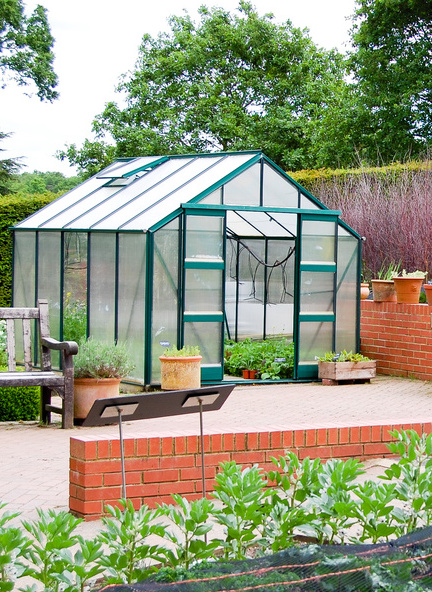}\hfill
  \includegraphics[height=10.9ex]{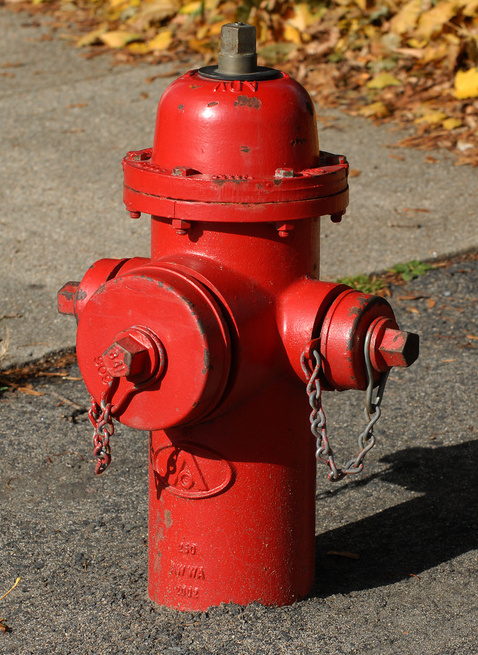}\hfill
  \includegraphics[height=10.9ex]{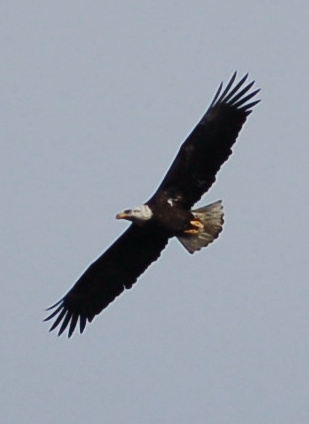}
  \includegraphics[height=10.9ex]{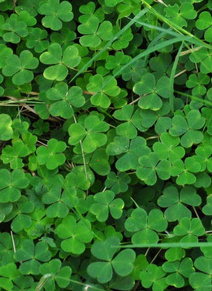}\hfill
  \includegraphics[height=10.9ex]{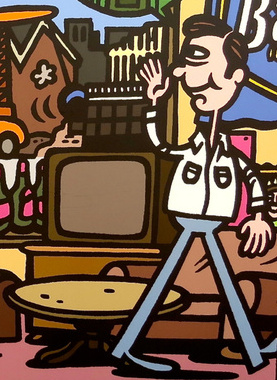}\hfill
  \includegraphics[height=10.9ex]{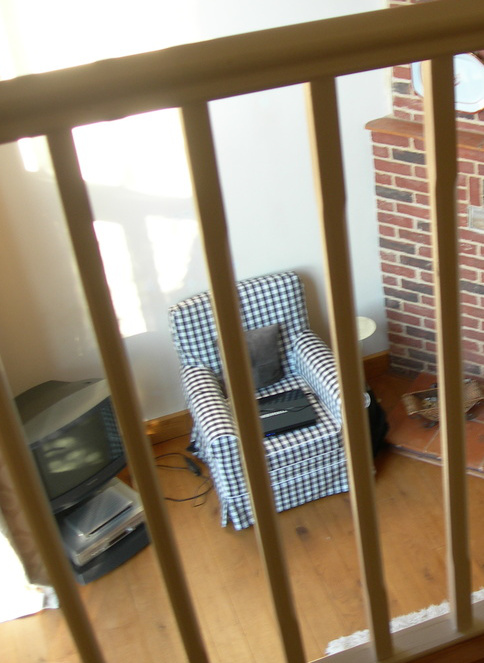}
\end{minipage}
\end{table}

\subsection{Segmentation and Instances}
\label{sec:dataset3}

Images were given to annotators for polygon segmentation of the entire image.
We instructed annotators to segment parts larger than a fingertip, ignore gaps smaller than a finger, and to follow material boundaries tightly while ignoring geometry and shadow boundaries. Following~\cite{opensurfaces}, annotators were instructed to segment glass and mirror surfaces rather than the covered or reflected surfaces. Unreal elements such as borders and watermarks were segmented separately. Images with objectionable content (\eg, violence) were not annotated.

\begin{figure}[t]
\includegraphics[height=12.1ex]{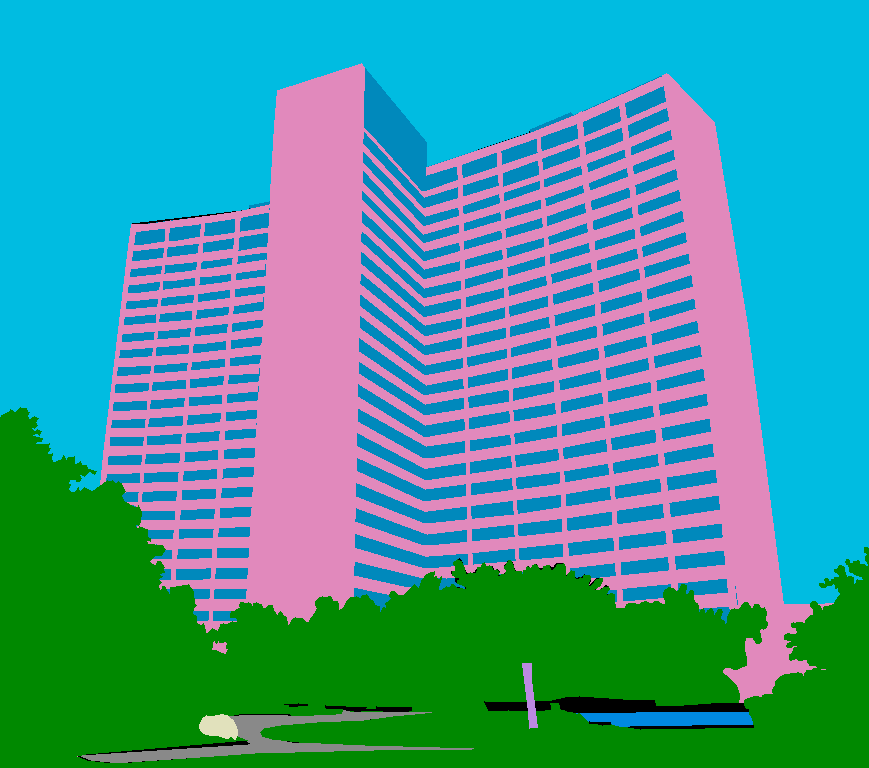}\hfill
\includegraphics[height=12.1ex]{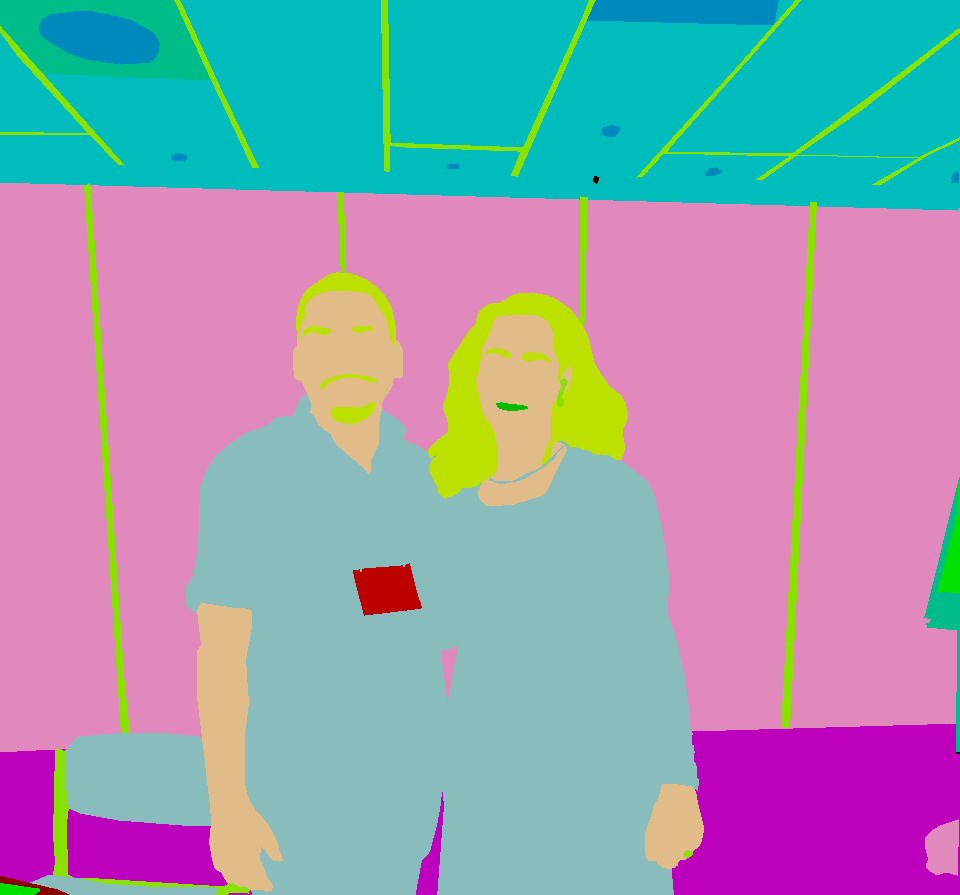}\hfill
\includegraphics[height=12.1ex]{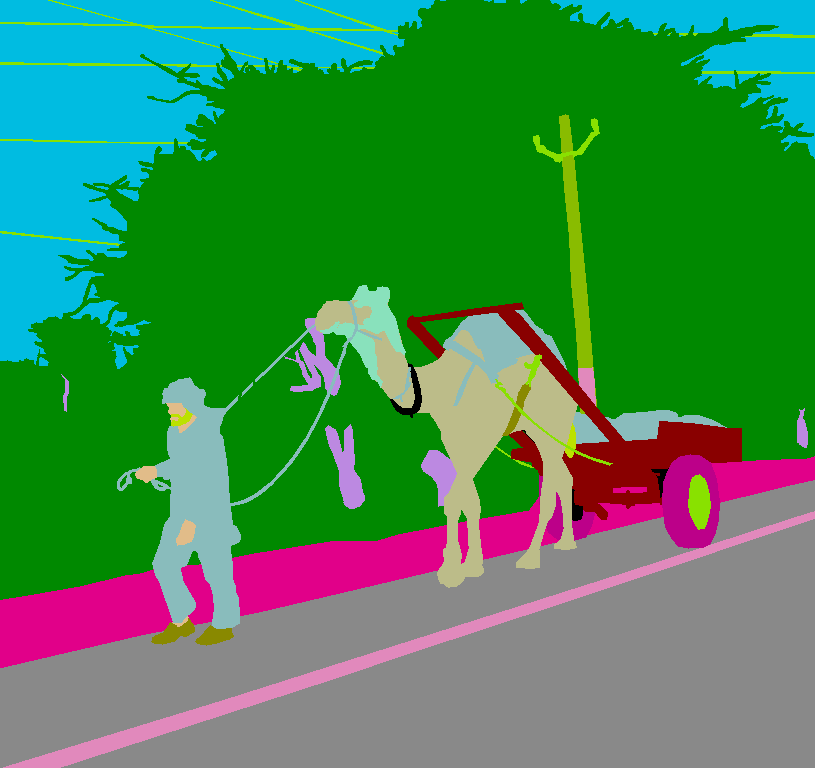}\hfill
\includegraphics[height=12.1ex]{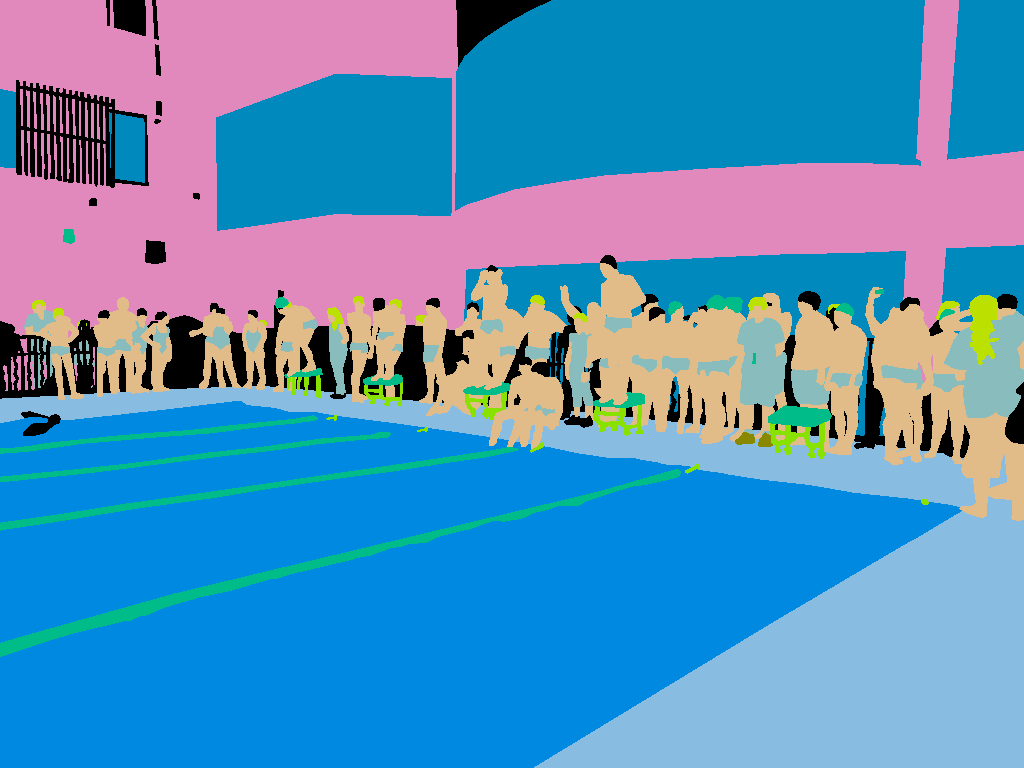}\hfill
\includegraphics[height=12.1ex]{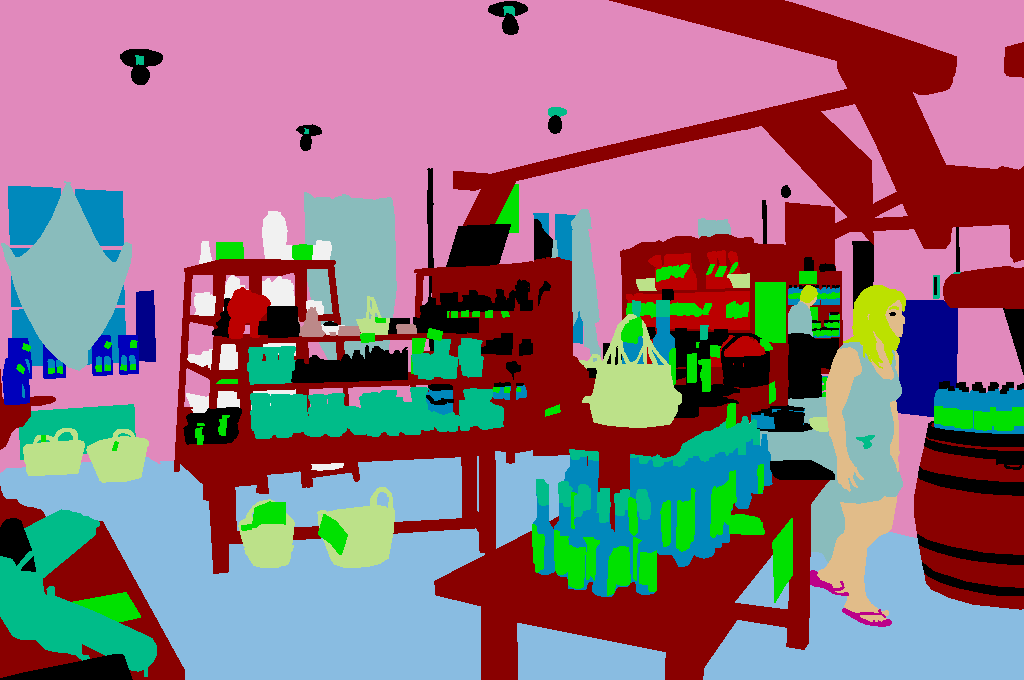}\hfill
\includegraphics[height=12.1ex]{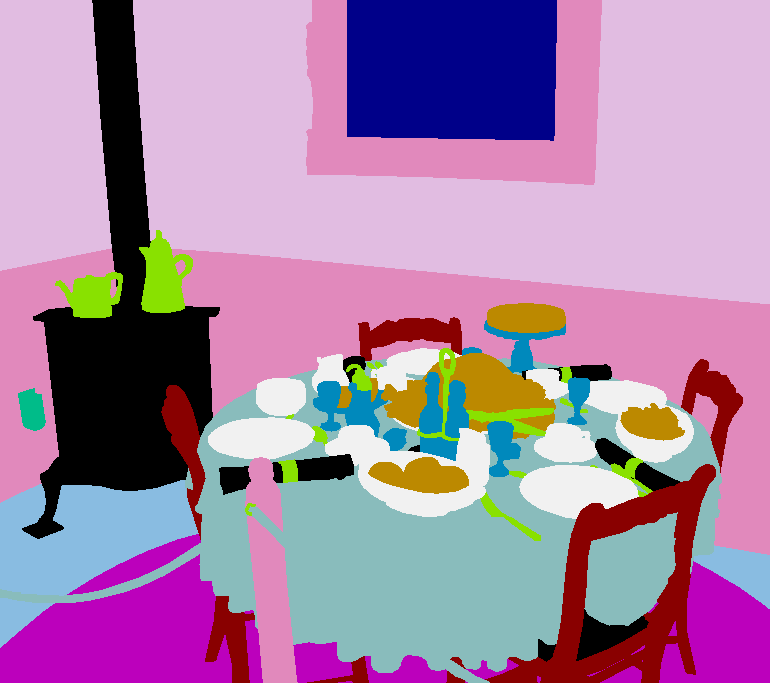}
\caption{{\bf Fused labels.} We show segmentation quality and variety of scenes, activities and materials ({\it left to right:} building exterior, workplace, road, swimming pool, shop, dining room). See Table~\ref{tab:fusedcount} for color legend. Black pixels are unlabeled (no consensus).}
\label{fig:fusedlabels}
\end{figure}

Annotators segmented resized images, with median longest edge of 1024 pixels, creating over 3.2 million segments (counting only those larger than 100 pixels) with a mean of 72 segments per image. The created segments are detailed---wires, jewelry, teeth, eyebrows, shoe soles, wheel rims, door hinges, clasps, buttons and latches are some of the small and thin materials segmented separately. See Figure~\ref{fig:teaser} and Figure~\ref{fig:fusedlabels} for examples of detailed segmentations.

We defined a material instance as materials of the same type from the same manufacturing source. For example a wooden cabinet should be segmented separately from a wood floor but the planks making up a single-source floor would be one instance. \OM is the first large-scale densely segmented dataset to have detailed material instances.

\begin{table}[t]
\centering
\caption{{\bf Annotator agreement rates.} High rates indicate consistent label assignment. Low rates indicate disagreement, confusion or unstructured error.}
\label{tab:agreement}
\begin{tabular}{@{}llp{3mm}llp{3mm}llp{3mm}ll@{}}\toprule
 Hair         & 0.95 & & Glass       & 0.80 & & Wood          & 0.67 & & Non-clear plastic & 0.60\\
 Skin         & 0.93 & & Paper       & 0.76 & & Tree wood     & 0.66 & & Leather           & 0.53\\
 Foliage      & 0.86 & & Carpet/rug  & 0.73 & & Tile          & 0.66 & & Cardboard         & 0.53\\
 Sky          & 0.86 & & Nat. stone  & 0.72 & & Metal         & 0.65 & & Artwork           & 0.51\\
 Food         & 0.84 & & Ceramic     & 0.70 & & Paint/plaster & 0.62 & & Clear plastic     & 0.50\\
 Fabric/cloth & 0.82 & & Mirror      & 0.68 & & Rubber        & 0.61 & & Concrete          & 0.45\\
\bottomrule
\end{tabular}
\end{table}

\subsection{Labeling}

The annotator who segmented an image also assigned labels based on their judgment and our instruction. We found that surfaces coated with another material or colored by absorbing ink required clarification. Appearance-changing coatings were labeled \matlabel{paint} while clear or appearance-enhancing coatings (\eg, varnish, cosmetics, sheer hosiery) were labeled as the underlying material. Small amounts of ink (\eg, printed text) are disregarded. Some surfaces imitate the appearance of other materials (\eg, laminate). High-quality imitations were labeled as the imitated material and low-quality imitations as the real material.

Our instructions were refined in each iteration and incorrect labels from early iterations were corrected. Some cases needed special instruction. We instructed annotators to label electronic displays as \matlabel{glass} and vinyl projection screens as \matlabel{not on list}. Uncovered artwork or photographs were to be labeled \matlabel{artwork} while glass-covered art should be labeled \matlabel{glass}. In ambiguous cases, we assume framed artwork has a glass cover.
\matlabel{Sky} includes day sky, night sky and aerial phenomenon (\eg, clouds, stars, moon, and sun).

We collected more opinions by presenting a segmentation, after removing labels, to a different annotator who relabeled the segments. The relabeling annotator could fix bad segments by adjusting polygons or assign special labels to indicate a segment does not follow boundaries or is made of multiple material types. We collected 98,526 opinions across 44,560 images consisting of 8.2 million segment labels (counting only segments larger than 100 pixels).

We studied label agreement by counting occurrences of a segment label and matching pixel-wise dominant label by a different annotator.
We found an agreement rate of 0.675. In cases of agreement, 8.9\% were unrecognizable (\matlabel{I cannot tell}) and 0.6\% were \matlabel{not on list}.
Table~\ref{tab:agreement} shows the agreement rate for classes larger than the median number of segments per class. Among the largest classes the most agreed-upon labels are \matlabel{hair}, \matlabel{skin}, \matlabel{foliage}, \matlabel{sky}, and \matlabel{food}.
We only analyze the largest classes since unstructured error (\eg, misclicks) can overwhelm the statistics of small classes, which are up to 2,720 times smaller.

\begin{table}[t]
\centering
\caption{{\bf Material occurrence in images.} We report the number of images in which a label occurs. The colors are used for visualizations.}
\label{tab:fusedcount}
\begin{tabular}{@{}p{3.1mm}lrp{3mm}p{3.1mm}lrp{3mm}p{3.1mm}lr@{}}\toprule
\cellcolor{matcolor29} & Paint/plaster & 39,323 & & \cellcolor{matcolor38} & Sky           & 3,306 & & \cellcolor{matcolor8}  & Chalkboard & 668 \\
\cellcolor{matcolor13} & Fabric/cloth  & 31,489 & & \cellcolor{matcolor27} & Mirror        & 3,242 & & \cellcolor{matcolor56}             & Asphalt    & 474 \\
\cellcolor{matcolor34} & Non-clear plas& 30,506 & & \cellcolor{matcolor4}  & Cardboard     & 3,150 & & \cellcolor{matcolor15}             & Fire       & 412 \\
\cellcolor{matcolor26} & Metal         & 30,504 & & \cellcolor{matcolor17} & Food          & 2,908 & & \cellcolor{matcolor19}             & Gemstone   & 369 \\
\cellcolor{matcolor20} & Glass         & 28,934 & & \cellcolor{matcolor10} & Concrete      & 2,853 & & \cellcolor{matcolor42}             & Sponge     & 326 \\
\cellcolor{matcolor52} & Wood          & 24,248 & & \cellcolor{matcolor6}  & Ceiling tile  & 2,524 & & \cellcolor{matcolor12}             & Eng. stone & 299 \\
\cellcolor{matcolor30} & Paper         & 20,763 & & \cellcolor{matcolor43} & Natural stone & 2,076 & & \cellcolor{matcolor25}             & Liquid     & 294 \\
\cellcolor{matcolor37} & Skin          & 18,524 & & \cellcolor{matcolor48} & Water         & 2,063 & & \cellcolor{matcolor31}             & Pearl      & 282 \\
\cellcolor{matcolor21} & Hair          & 17,766 & & \cellcolor{matcolor53} & Tree wood     & 2,026 & & \cellcolor{matcolor11}             & Cork       & 273 \\
\cellcolor{matcolor16} & Foliage       & 11,384 & & \cellcolor{matcolor51} & Wicker        & 1,895 & & \cellcolor{matcolor36}             & Sand       & 272 \\
\cellcolor{matcolor46} & Tile          & 10,173 & & \cellcolor{matcolor41} & Soil/mud      & 1,855 & & \cellcolor{matcolor39}             & Snow       & 191 \\
\cellcolor{matcolor5}  & Carpet/rug    &  9,516 & & \cellcolor{matcolor44} & Pol. stone    & 1,831 & & \cellcolor{matcolor40}             & Soap       & 154 \\
\cellcolor{matcolor7}  & Ceramic       &  8,314 & & \cellcolor{matcolor3}  & Brickwork     & 1,654 & & \cellcolor{matcolor9}              & Clutter    & 128 \\
\cellcolor{matcolor35} & Rubber        &  7,811 & & \cellcolor{matcolor18} & Fur           & 1,567 & & \cellcolor{matcolor23}             & Ice        & 96 \\
\cellcolor{matcolor24} & Leather       &  7,354 & & \cellcolor{matcolor50} & Whiteboard    & 1,171 & & \cellcolor{matcolor45}             & Styrofoam  & 88 \\
\cellcolor{matcolor33} & Clear plastic &  6,431 & & \cellcolor{matcolor49} & Wax           & 1,107 & & \cellcolor{matcolor14}             & Fiberglass wool & 33 \\
\cellcolor{matcolor32} & Artwork       &  4,344 & & \cellcolor{matcolor47} & Wallpaper     & 1,076 \\
\cellcolor{matcolor2}  & Bone/horn     &  3,751 & & \cellcolor{matcolor1}  & Animal skin   & 1,007 \\
\bottomrule
\end{tabular}
\end{table}

\subsection{Label Fusion}

Each annotator's segments are rendered to create a label map. Label maps were inspected for correctness and we fixed incorrect labels in 1,803 images.
Next, we create a single \emph{fused label map} for each image. First, we combined label maps pixel-wise by taking the strict majority label. Next, we overlaid manual corrections and reassigned non-semantic labels (\eg, \matlabel{I cannot tell}) to \matlabel{no label}.
The fused maps have a mean labeled area fraction of 0.784. For comparison, we created fused label maps for OpenSurfaces and found its density is 0.210. \OM is 2.3x larger and 3.7x denser, which is 8.4x more labeled area. 
Compared to the 3M points in MINC~\cite{minc}, DMS has 3.2M fused segments which carry more information about shape, boundary and co-occurrences. While MINC annotations span 10x more images, point annotations cannot evaluate segmentation boundaries for scene parsing tasks.
Example fused maps and class occurrences are shown in Figure~\ref{fig:fusedlabels} and Table~\ref{tab:fusedcount}.
The smallest class appears in 33 images whereas the largest class, \matlabel{paint}, appears in 39,323 images, which is 88\% of the images.

\captionsetup[subfigure]{labelformat=empty}

\section{Experiments}
\label{sec:experiments}

First, we investigate the impact of our data on training deep learning models with a cross-dataset comparison (Section~\ref{sec:crossdataset}). Then, we compare the impact of skin type distributions on fairness of skin recognition (Section~\ref{sec:skin}). Next, we establish a material segmentation benchmark for 46 kinds of materials (Section~\ref{sec:baseline}). Finally, we show predictions on real world images (Section~\ref{sec:real}).

{\bf Splits.} We created train, validation and test splits for our data by assigning images according to material occurrence. The smallest classes are assigned a ratio of 1\,:\,1\,:\,1, which increases to 2.5\,:\,1\,:\,1 for the largest. An image assignment impacts the ratio of multiple classes so small classes are assigned first. There are 24,255 training images, 10,139 validation images and 10,166 test images.

\subsection{Cross-Dataset Comparison}
\label{sec:crossdataset}

Does training with our data lead to a better model?
This experiment compares a model fit to our data against two baselines fit to OpenSurfaces data---the strongest published model~\cite{upernet} and a model with the same architecture as ours. 
There are two sources of data. The first is OpenSurfaces data with the splits and 25 labels proposed by~\cite{upernet}. The second is comparable DMS training and validation data (\cite{upernet} does not define a test split) created by translating our labels to match~\cite{upernet}. The evaluation set, which we call Avg-Val, is made of both parts---the validation sets of OpenSurfaces and \OM, called OS-Val and \OM-Val, respectively---weighted equally.
For evaluation of our data we fit models to DMS training data and choose the model that performs best on \OM-Val. This model, which we call \OM-25, is a ResNet-50 architecture~\cite{he2016deep} with dilated convolutions~\cite{chen2017deeplab,yu2015multi} as the encoder, and Pyramid Pooling Module from PSPNet~\cite{zhao2017pyramid} as the decoder.
The first baseline (Table~\ref{tab:results1}, row 2) is UPerNet~\cite{upernet}, a multitask scene parsing model which uses cross-domain knowledge to boost material segmentation performance. The second baseline (Table~\ref{tab:results1}, row 3), called OS-25,  has the same architecture as \OM-25 but is fit to OpenSurfaces training data.
Table~\ref{tab:results1} shows the results. We report per-pixel accuracy (Acc), mean class accuracy (mAcc), mean intersection-over-union (mIoU) and $\Delta$, the absolute difference in a metric across \OM-Val and OS-Val. A low $\Delta$ indicates a model is more consistent across datasets.
We find that fitting a model to \OM training data leads to higher performance and lower $\Delta$ on all metrics.
We also report the metrics on each validation set and find that both baselines underperform on \OM-Val.
We find that DMS-25 performs 0.01 lower on OS-Val mAcc compared to a model trained on OpenSurfaces data. This may be due to differences in annotation and image variety. We use our photographic type labels to investigate the larger performance gaps on \OM-Val.

Why do models trained with OpenSurfaces underperform on our validation images?
In Table~\ref{tab:results2} we report per-pixel accuracy of \OM-25, UPerNet, and OS-25 across nine categories. We find that \OM-25 performs consistently across categories with the lowest performing category (unreal images) 0.071 below the highest performing category (images of enclosed areas). UPerNet shows lower performance across all categories with a drop of 0.426 from images of enclosed areas to images of distant outdoor things. And OS-25 shows similar performance with a drop of 0.407. We observe that both UPerNet and OS-25 have low performance on outdoor images and images without any context.
This study shows that photographic types can improve our understanding of how material segmentation models perform in different settings. And, these results justify our decision to collect outdoor images and images of different photographic types.

\begin{table}[t]
\centering
\caption{{\bf Training data evaluation.} We compare segmentation of 25 materials with our training data ({\it row 1}) to OpenSurfaces data with two kinds of models ({\it rows 2 and 3}). Avg-Val is the equally-weighted validation sets of each dataset, \OM-Val and OS-Val. $\Delta$ is the difference in a metric across datasets. A convnet fit to our data achieves higher performance and is more consistent across datasets.}
\label{tab:results1}
\begin{tabular}{@{}lrlrlrcccccc@{}}\toprule
 Training data                    & &  Model                   & & Metric & & {Avg-Val }$\uparrow$ & $\Delta\downarrow$ & {\OM-Val }$\uparrow$  & {OS-Val }$\uparrow$\\\midrule
                                  & &                          & & Acc    & & {\bf 0.777}          & {\bf 0.047}        & 0.753                 & 0.800\\
 \OM (Ours)                       & &  DMS-25                  & & mAcc   & & {\bf 0.689}          & {\bf 0.006}        & 0.686                 & 0.692\\
                                  & &                          & & mIoU   & & {\bf 0.500}          & {\bf 0.014}        & 0.507                 & 0.493\\\midrule
                                  & &                          & & Acc    & & 0.682                &  0.310             & 0.527                 & 0.837\\
 OpenSurfaces~\cite{opensurfaces} & &  UPerNet~\cite{upernet}  & & mAcc   & & 0.486                &  0.274             & 0.349                 & 0.623\\
                                  & &                          & & mIoU   & & 0.379                &  0.298             & 0.230                 & 0.528\\\midrule
                                  & &                          & & Acc    & & 0.705                &  0.231             & 0.589                 & 0.820\\
 OpenSurfaces~\cite{opensurfaces} & &  OS-25                   & & mAcc   & & 0.606                &  0.193             & 0.509                 & 0.702\\
                                  & &                          & & mIoU   & & 0.416                &  0.199             & 0.316                 & 0.515\\
\bottomrule
\end{tabular}
\end{table}

\begin{table}[t]
\centering
\caption{{\bf Performance analysis with photographic types.} A model fit to our data, \OM-25 ({\it Table~\ref{tab:results1}, row 1}), performs well on all photographic types whereas two models fit to OpenSurfaces, UPerNet and OS-25 ({\it Table~\ref{tab:results1}, rows 2-3}) have low performance outdoors ({\it middle}) and on surfaces without any context ({\it row 7}).}
\label{tab:results2}
\begin{tabular}{@{}lccccc@{}}\toprule
 \multicolumn{1}{l}{Photographic Type}           & \multicolumn{5}{c}{Per-Pixel Accuracy}\\
 \cmidrule{2-6}
                                                 & \OM-25 (Ours)  & & UPerNet~\cite{upernet} & & OS-25\\\midrule
 An area with visible enclosure                  & 0.756  & & 0.615 & & 0.632\\
 A collection of indoor things                   & 0.752  & & 0.546 & & 0.622\\
 A tightly cropped indoor thing                  & 0.710  & & 0.441 & & 0.561\\\midrule
 A view of reachable outdoor things              & 0.750  & & 0.265 & & 0.388\\
 A tightly cropped outdoor thing                 & 0.731  & & 0.221 & & 0.359\\
 Distant unreachable outdoor things              & 0.736  & & 0.189 & & 0.225\\\midrule
 A real surface without context                  & 0.691  & & 0.222 & & 0.348\\
 Not a real photo                                & 0.685  & & 0.528 & & 0.551\\
 An obstructed or distorted view                 & 0.729  & & 0.370 & & 0.496\\
\bottomrule
\end{tabular}
\end{table}

\begin{table}[t]
\centering
\caption{{\bf Test set results.} We report metrics for our model, \OM-46. 17 materials, in italics, are new---not predicted by prior general-purpose models~\cite{minc,upernet,schwartz2019recognizing}.}
\label{tab:OM46_per_class_testset}
\begin{tabular}{@{}lccp{3mm}lccp{3mm}lcc@{}}\toprule
Category               &   Acc & IoU   & & Category                   & Acc   & IoU   & & Category            & Acc & IoU\\\midrule
Sky                    & 0.962 & 0.892 & &  \textit{Chalkboard}       & 0.712 & 0.548 & & \textit{Artwork}    & 0.454 & 0.301 \\
Fur                    & 0.910 & 0.707 & & Paint/plaster              & 0.694 & 0.632 & & Mirror & 0.452      & 0.278 \\
Foliage                & 0.902 & 0.761 & & Wicker                     & 0.674 & 0.460 & & \textit{Sand}       & 0.444 & 0.340 \\
Skin                   & 0.886 & 0.640 & & Natural stone              & 0.665 & 0.436 & & \textit{Ice}        & 0.440 & 0.362 \\
Hair                   & 0.881 & 0.673 & & Glass                      & 0.653 & 0.483 & & \textit{Tree wood}  & 0.428 & 0.261 \\
Food                   & 0.868 & 0.668 & & Asphalt                    & 0.628 & 0.442 & & Pol. stone          & 0.379 & 0.236 \\
 \textit{Ceiling tile} & 0.867 & 0.611 & & Leather                    & 0.615 & 0.373 & & \textit{Clear plastic} & 0.360 & 0.222 \\
Water                  & 0.866 & 0.712 & & \textit{Snow}              & 0.610 & 0.465 & & Rubber              & 0.255  & 0.163 \\
Carpet/rug             & 0.849 & 0.592 & & Concrete                   & 0.603 & 0.304 & & \textit{Clutter}    & 0.182 & 0.152 \\
 \textit{Whiteboard}   & 0.838 & 0.506 & & Metal                      & 0.575 & 0.303 & & \textit{Fire}       & 0.176 & 0.147 \\
Fabric/cloth           & 0.801 & 0.692 & & \textit{Wax}               & 0.573 & 0.371 & & \textit{Gemstone}   & 0.116 & 0.096 \\
Wood                   & 0.797 & 0.635 & & Cardboard                  & 0.570 & 0.363 & & Eng. stone          & 0.088 & 0.071 \\
Ceramic                & 0.757 & 0.427 & & Wallpaper                  & 0.544 & 0.329 & & \textit{Cork}       & 0.082 & 0.066 \\
Brickwork              & 0.746 & 0.491 & & \textit{Non-clear plastic} & 0.519 & 0.321 & & \textit{Bone/horn}  & 0.074 & 0.070 \\
Paper                  & 0.729 & 0.508 & & Soil/mud                   & 0.511 & 0.332 \\
Tile                   & 0.722 & 0.550 & & \textit{Animal skin}       & 0.472 & 0.308 \\
\bottomrule
\end{tabular}
\end{table}

\subsection{Recognition of Different Skin Types}
\label{sec:skin}
Models trained on face datasets composed of unbalanced skin types exhibit classification disparities~\cite{gendershades}. Does this impact skin recognition? Without any corrections for skin type imbalance we find
that \OM-25 has a 3\% accuracy gap among different skin types on \OM-val (Type I-II: 0.933, Type III-IV: 0.924, Type V-VI: 0.903) while OS-25 has a larger gap of 13.3\% (Type I-II: 0.627, Type III-IV: 0.571, Type V-VI: 0.494). This confirms that skin type imbalance impacts skin recognition. Our contribution lies in providing more data for all skin types (Table~\ref{tab:skintypes}), which makes it easier for practitioners to create fair models.

\subsection{A Material Segmentation Benchmark}
\label{sec:baseline}
It is common practice to select large categories and combine smaller ones (our smallest occurs in only 12 training images) for a benchmark. Yet, we cannot know {\it a priori} how much training data is sufficient to learn a category. We choose to be guided by the validation data. We fit many models to all 52 categories then inspect the results to determine which categories can be reliably learned.
We select ResNet50~\cite{he2016deep} with dilated convolutions~\cite{chen2017deeplab,yu2015multi} as the encoder, and Pyramid Pooling Module from PSPNet~\cite{zhao2017pyramid} as the decoder. We choose this architecture because it has been shown to be effective for scene parsing~\cite{zhao2017pyramid,zhou2019semantic}.
Our best model, which we call \OM-52, predicts 52 materials with per-pixel accuracy 0.735, mean class accuracy 0.535 and mIoU 0.392 on \OM-val.

We inspected a few strongest \OM-52 fitted models and found that 6 categories consistently stood out as underperforming---having 0 accuracy in some cases and, at best, not much higher than chance. Those categories are \matlabel{non-water liquid}, \matlabel{fiberglass}, \matlabel{sponge}, \matlabel{pearl}, \matlabel{soap} and \matlabel{styrofoam}, which occur in 129, 12, 149, 129, 58 and 33 training images, respectively. Guided by this discovery we select the other 46 material labels for a benchmark.

We train a model, called \OM-46, to predict the selected categories, with the same architecture as DMS-52. We use a batch size of 64 and stochastic gradient descent optimizer with 1e-3 base learning rate and 1e-4 weight decay. We use ImageNet pretraining~\cite{zhou2017scene,zhou2019semantic} to initialize the encoder weights, and scale the learning rate for the encoder by 0.25. We update the learning rate with a cosine annealing schedule with warm restart~\cite{loshchilov2016sgdr} every 30 epochs for 60 epochs. Because the classes are imbalanced we use weighted symmetric cross entropy~\cite{wang2019symmetric}, computed across \OM training images, as the loss function, which gives more weight to classes with fewer ground truth pixels. We apply stochastic transformations for data augmentation (scale, horizontal and vertical flips, color jitter, Gaussian noise, Gaussian blur, rotation and crop), scale inputs into [0, 1], and normalize with mean = [0.485, 0.456, 0.406] and std = [0.229, 0.224, 0.225] from ImageNet~\cite{deng2009imagenet}. The training tensor has height and width of 512.

\OM-46 predicts 46 materials with per-pixel accuracy 0.731/0.729, mean class accuracy 0.598/0.585 and mIoU 0.435/0.420 on \OM-val/\OM-test respectively. We report the test set per-class accuracy and IoU in Table~\ref{tab:OM46_per_class_testset}. We find that \matlabel{sky}, \matlabel{fur}, \matlabel{foliage}, \matlabel{skin} and \matlabel{hair} have the highest recognition rates, similar to the findings of~\cite{minc}. 17 materials do not appear in any prior large-scale material benchmarks. Among these new materials we report high recognition rates for \matlabel{ceiling tile}, \matlabel{whiteboard} and \matlabel{chalkboard}.
To our knowledge, \OM-46 is the first material segmentation model evaluated on large-scale dense segmentations and predicts more classes than any general-purpose model.

\begin{figure}[t]
\includegraphics[height=13.0ex]{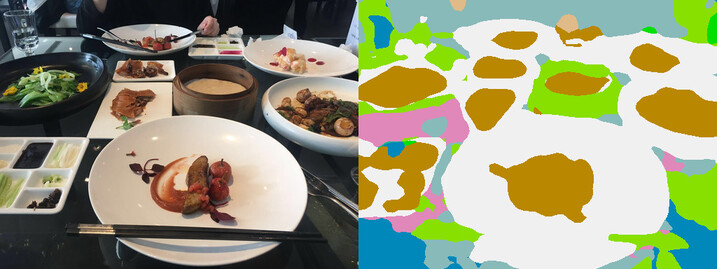}\hfill
\includegraphics[height=13.0ex]{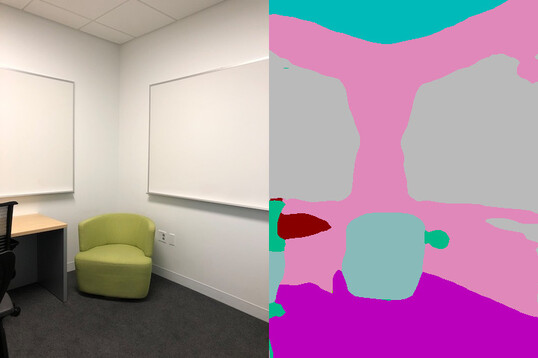}\hfill
\includegraphics[height=13.0ex]{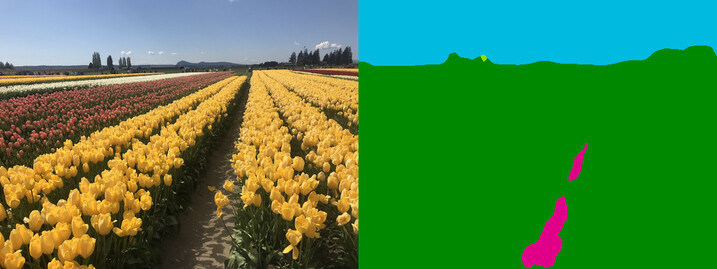}
\includegraphics[height=14.4ex]{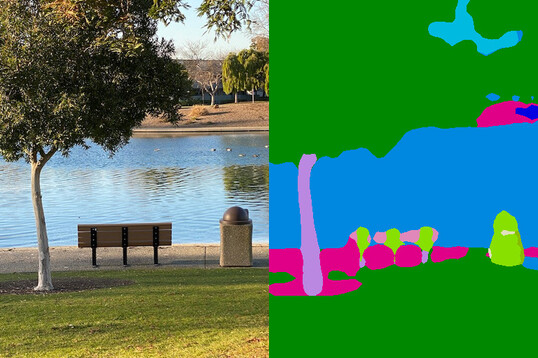}\hfill
\includegraphics[height=14.4ex]{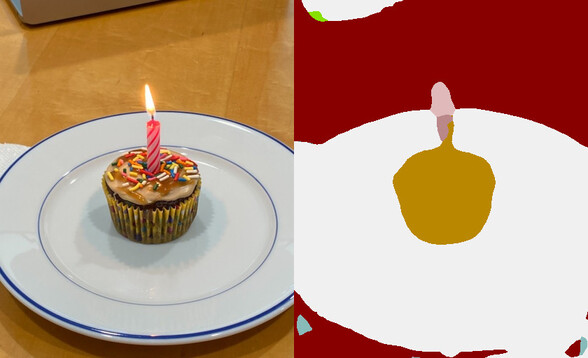}\hfill
\includegraphics[height=14.4ex]{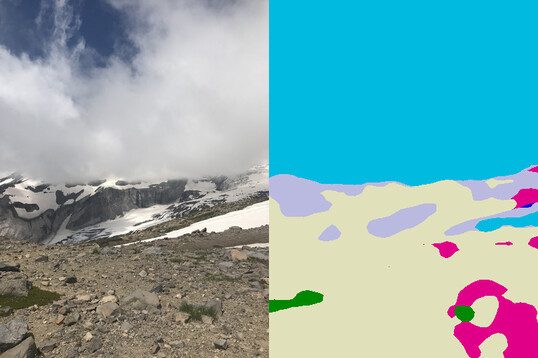}\hfill
\includegraphics[height=14.4ex]{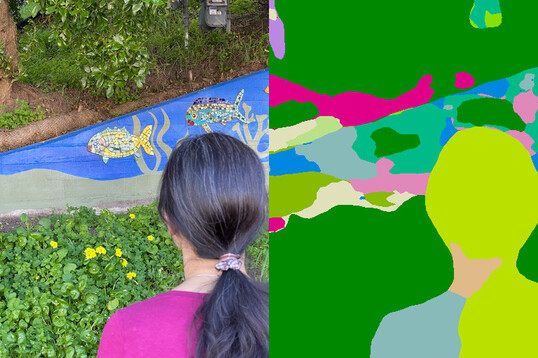}

\caption{{\bf Real-world examples.} Our model, \OM-46, predicts 46 kinds of indoor and outdoor materials. See Table~\ref{tab:fusedcount} for color legend.}
\label{fig:examples}
\end{figure}

\subsection{Real-World Examples}
\label{sec:real}
In Figure~\ref{fig:examples} we demonstrate \OM-46 on indoor and outdoor photos from daily life. Our model recognizes and localizes \matlabel{food} on \matlabel{ceramic} plates, workplace materials (\matlabel{whiteboard} and \matlabel{ceiling tile}), ground cover materials (\matlabel{soil}, \matlabel{stone}, \matlabel{foliage} and \matlabel{snow}), unprocessed \matlabel{tree wood}, and \matlabel{fire} on a \matlabel{wax} candle.

{\bf A Failure Case.} 
The last image is a failure case where our model is confused by decorative tile artwork. We also see opportunities for further improving boundaries and localizing small surfaces.

\section{Discussion and Conclusion}
\label{sec:conclusion}

{\bf Dense Annotation.}
Prior works~\cite{opensurfaces,minc,schwartz2019recognizing} instruct annotators to locate and segment regions made of a given material. Our approach is different. We instruct annotators to segment and label the entire image.
This approach collects different data because annotators address all surfaces---not just those which are readily recognized. We hypothesize this creates a more difficult dataset, and propose this approach is necessary for evaluation of scene parsing, which predicts all pixels.

{\bf Real vs. Synthetic.}
Synthetic data has achieved high levels of realism (\eg, Hypersim~\cite{hypersim}) and may be a valuable generator of training data. We opted to label real photos because models trained on synthetic data need a real evaluation dataset to confirm the domain gap from synthetic to real has been bridged.

{\bf Privacy.}
Material predictions can be personal. Knowing a limb is not made of skin reveals a prosthetic. The amount of body hair reveals one aspect of appearance. Precious materials in a home reveals socio-economic status. Clothing material indicates degree of nakedness.
Care is needed if material segmentation is tied to identity.
Limiting predicted materials to only those needed by an application or separating personal materials from identity are two ways, among many possible ways, to strengthen privacy and protect personal information.

\section{Conclusion}

We present the first large-scale densely-annotated material segmentation dataset which can train or evaluate indoor and outdoor scene parsing models.
\footnote{Our data is available at https://github.com/apple/ml-dms-dataset.}
We propose a benchmark on 46 kinds of materials. Our data can be a foundation for algorithms which utilize material type, make use of physical properties for simulations or functional properties for planning and human-computer interactions. We look forward to expanding the number of materials, finding new methods to reach even better full-scene material segmentation, and combining the point-wise annotations of MINC~\cite{minc} with our data in future work.

{\bf Acknowledgements.}
We thank Allison Vanderby, Hillary Strickland, Laura Snarr, Mya Exum, Subhash Sudan, Sneha Deshpande, and Doris Guo for their help with acquiring data; Richard Gass, Daniel Kurz and Selim Ben Himane for their support.

\par\vfill\par
\bibliographystyle{splncs04}
\bibliography{bib}

\begin{thebibliography}{10}
\providecommand{\url}[1]{\texttt{#1}}
\providecommand{\urlprefix}{URL }
\providecommand{\doi}[1]{https://doi.org/#1}

\bibitem{thingsstuff}
Adelson, E.H.: On seeing stuff: The perception of materials by humans and
  machines. In: Human vision and electronic imaging VI. vol.~4299, pp. 1--12.
  SPIE (2001)

\bibitem{opensurfaces}
Bell, S., Upchurch, P., Snavely, N., Bala, K.: Open{S}urfaces: A richly
  annotated catalog of surface appearance. ACM Transactions on graphics (TOG)
  \textbf{32}(4),  1--17 (2013)

\bibitem{minc}
Bell, S., Upchurch, P., Snavely, N., Bala, K.: Material recognition in the wild
  with the {M}aterials in {C}ontext database. In: Proceedings of the IEEE
  conference on computer vision and pattern recognition. pp. 3479--3487 (2015)

\bibitem{brandao2016material}
Brandao, M., Shiguematsu, Y.M., Hashimoto, K., Takanishi, A.: Material
  recognition {CNN}s and hierarchical planning for biped robot locomotion on
  slippery terrain. In: 2016 IEEE-RAS 16th International Conference on Humanoid
  Robots (Humanoids). pp. 81--88. IEEE (2016)

\bibitem{gendershades}
Buolamwini, J., Gebru, T.: Gender shades: Intersectional accuracy disparities
  in commercial gender classification. In: Conference on fairness,
  accountability and transparency. pp. 77--91. PMLR (2018)

\bibitem{cocostuff}
Caesar, H., Uijlings, J., Ferrari, V.: {COCO}-{S}tuff: Thing and stuff classes
  in context. In: Proceedings of the IEEE conference on computer vision and
  pattern recognition. pp. 1209--1218 (2018)

\bibitem{chen2017deeplab}
Chen, L.C., Papandreou, G., Kokkinos, I., Murphy, K., Yuille, A.L.: Deep{L}ab:
  Semantic image segmentation with deep convolutional nets, atrous convolution,
  and fully connected {CRF}s. IEEE transactions on pattern analysis and machine
  intelligence  \textbf{40}(4),  834--848 (2017)

\bibitem{chen2020context}
Chen, L., Tang, W., John, N.W., Wan, T.R., Zhang, J.J.: Context-aware mixed
  reality: A learning-based framework for semantic-level interaction. In:
  Computer Graphics Forum. vol.~39, pp. 484--496. Wiley Online Library (2020)

\bibitem{deng2009imagenet}
Deng, J., Dong, W., Socher, R., Li, L.J., Li, K., Fei-Fei, L.: Imagenet: A
  large-scale hierarchical image database. In: 2009 IEEE conference on computer
  vision and pattern recognition. pp. 248--255. Ieee (2009)

\bibitem{fitzpatrick}
Fitzpatrick, T.B.: The validity and practicality of sun-reactive skin types {I}
  through {VI}. Archives of dermatology  \textbf{124}(6),  869--871 (1988)

\bibitem{gao2016deep}
Gao, Y., Hendricks, L.A., Kuchenbecker, K.J., Darrell, T.: Deep learning for
  tactile understanding from visual and haptic data. In: 2016 IEEE
  International Conference on Robotics and Automation (ICRA). pp. 536--543.
  IEEE (2016)

\bibitem{detectron}
Girshick, R., Radosavovic, I., Gkioxari, G., Doll\'{a}r, P., He, K.: Detectron.
  \url{https://github.com/facebookresearch/detectron} (2018)

\bibitem{he2016deep}
He, K., Zhang, X., Ren, S., Sun, J.: Deep residual learning for image
  recognition. In: Proceedings of the IEEE conference on computer vision and
  pattern recognition. pp. 770--778 (2016)

\bibitem{hu2011toward}
Hu, D., Bo, L., Ren, X.: Toward robust material recognition for everyday
  objects. In: BMVC. vol.~2, p.~6. Citeseer (2011)

\bibitem{fashionpedia}
Jia, M., Shi, M., Sirotenko, M., Cui, Y., Cardie, C., Hariharan, B., Adam, H.,
  Belongie, S.: Fashionpedia: Ontology, segmentation, and an attribute
  localization dataset. In: European conference on computer vision. pp.
  316--332. Springer (2020)

\bibitem{dlib}
King, D.E.: Dlib-ml: A machine learning toolkit. The Journal of Machine
  Learning Research  \textbf{10},  1755--1758 (2009)

\bibitem{openimages2}
Krasin, I., Duerig, T., Alldrin, N., Ferrari, V., Abu-El-Haija, S., Kuznetsova,
  A., Rom, H., Uijlings, J., Popov, S., Kamali, S., Malloci, M., Pont-Tuset,
  J., Veit, A., Belongie, S., Gomes, V., Gupta, A., Sun, C., Chechik, G., Cai,
  D., Feng, Z., Narayanan, D., Murphy, K.: Open{I}mages: A public dataset for
  large-scale multi-label and multi-class image classification. Dataset
  available from https://storage.googleapis.com/openimages/web/index.html
  (2017)

\bibitem{coco}
Lin, T.Y., Maire, M., Belongie, S., Hays, J., Perona, P., Ramanan, D.,
  Doll{\'a}r, P., Zitnick, C.L.: Microsoft {COCO}: Common objects in context.
  In: European conference on computer vision. pp. 740--755. Springer (2014)

\bibitem{loshchilov2016sgdr}
Loshchilov, I., Hutter, F.: {SGDR}: Stochastic gradient descent with warm
  restarts. In: International Conference on Learning Representations (2017)

\bibitem{mei2020don}
Mei, H., Yang, X., Wang, Y., Liu, Y., He, S., Zhang, Q., Wei, X., Lau, R.W.:
  Don't hit me! glass detection in real-world scenes. In: Proceedings of the
  IEEE/CVF Conference on Computer Vision and Pattern Recognition. pp.
  3687--3696 (2020)

\bibitem{murmann2019dataset}
Murmann, L., Gharbi, M., Aittala, M., Durand, F.: A dataset of
  multi-illumination images in the wild. In: Proceedings of the IEEE/CVF
  International Conference on Computer Vision. pp. 4080--4089 (2019)

\bibitem{ordonez2013large}
Ordonez, V., Deng, J., Choi, Y., Berg, A.C., Berg, T.L.: From large scale image
  categorization to entry-level categories. In: Proceedings of the ieee
  international conference on computer vision. pp. 2768--2775 (2013)

\bibitem{park2018photoshape}
Park, K., Rematas, K., Farhadi, A., Seitz, S.M.: Photo{S}hape: Photorealistic
  materials for large-scale shape collections. ACM Trans. Graph.
  \textbf{37}(6) (Nov 2018)

\bibitem{sunattributes}
Patterson, G., Hays, J.: {SUN} attribute database: Discovering, annotating, and
  recognizing scene attributes. In: 2012 IEEE Conference on Computer Vision and
  Pattern Recognition. pp. 2751--2758. IEEE (2012)

\bibitem{ritchie2021material}
Ritchie, J.B., Paulun, V.C., Storrs, K.R., Fleming, R.W.: Material perception
  for philosophers. Philosophy Compass  \textbf{16}(10),  e12777 (2021)

\bibitem{hypersim}
Roberts, M., Ramapuram, J., Ranjan, A., Kumar, A., Bautista, M.A., Paczan, N.,
  Webb, R., Susskind, J.M.: Hypersim: A photorealistic synthetic dataset for
  holistic indoor scene understanding. In: Proceedings of the IEEE/CVF
  International Conference on Computer Vision. pp. 10912--10922 (2021)

\bibitem{labelme}
Russell, B.C., Torralba, A., Murphy, K.P., Freeman, W.T.: Label{M}e: A database
  and web-based tool for image annotation. International journal of computer
  vision  \textbf{77}(1),  157--173 (2008)

\bibitem{cleargrasp}
Sajjan, S., Moore, M., Pan, M., Nagaraja, G., Lee, J., Zeng, A., Song, S.:
  Clear{G}rasp: 3{D} shape estimation of transparent objects for manipulation.
  In: 2020 IEEE International Conference on Robotics and Automation (ICRA). pp.
  3634--3642. IEEE (2020)

\bibitem{schissler2017acoustic}
Schissler, C., Loftin, C., Manocha, D.: Acoustic classification and
  optimization for multi-modal rendering of real-world scenes. IEEE
  transactions on visualization and computer graphics  \textbf{24}(3),
  1246--1259 (2017)

\bibitem{schwartz2019recognizing}
Schwartz, G., Nishino, K.: Recognizing material properties from images. IEEE
  transactions on pattern analysis and machine intelligence  \textbf{42}(8),
  1981--1995 (2019)

\bibitem{sharan2013recognizing}
Sharan, L., Liu, C., Rosenholtz, R., Adelson, E.H.: Recognizing materials using
  perceptually inspired features. International journal of computer vision
  \textbf{103}(3),  348--371 (2013)

\bibitem{fmd}
Sharan, L., Rosenholtz, R., Adelson, E.H.: Accuracy and speed of material
  categorization in real-world images. Journal of vision  \textbf{14}(9),
  12--12 (2014)

\bibitem{figaro}
Svanera, M., Muhammad, U.R., Leonardi, R., Benini, S.: Figaro, hair detection
  and segmentation in the wild. In: 2016 IEEE International Conference on Image
  Processing (ICIP). pp. 933--937. IEEE (2016)

\bibitem{van2021materials}
Van~Zuijlen, M.J., Lin, H., Bala, K., Pont, S.C., Wijntjes, M.W.: Materials in
  {P}aintings ({MIP}): An interdisciplinary dataset for perception, art
  history, and computer vision. Plos one  \textbf{16}(8),  e0255109 (2021)

\bibitem{wang20164d}
Wang, T.C., Zhu, J.Y., Hiroaki, E., Chandraker, M., Efros, A.A., Ramamoorthi,
  R.: A {4D} light-field dataset and {CNN} architectures for material
  recognition. In: European conference on computer vision. pp. 121--138.
  Springer (2016)

\bibitem{wang2019symmetric}
Wang, Y., Ma, X., Chen, Z., Luo, Y., Yi, J., Bailey, J.: Symmetric cross
  entropy for robust learning with noisy labels. In: Proceedings of the
  IEEE/CVF International Conference on Computer Vision. pp. 322--330 (2019)

\bibitem{upernet}
Xiao, T., Liu, Y., Zhou, B., Jiang, Y., Sun, J.: Unified perceptual parsing for
  scene understanding. In: Proceedings of the European Conference on Computer
  Vision (ECCV). pp. 418--434 (2018)

\bibitem{gtos2}
Xue, J., Zhang, H., Dana, K.: Deep texture manifold for ground terrain
  recognition. In: Proceedings of the IEEE Conference on Computer Vision and
  Pattern Recognition. pp. 558--567 (2018)

\bibitem{gtos}
Xue, J., Zhang, H., Dana, K., Nishino, K.: Differential angular imaging for
  material recognition. In: Proceedings of the IEEE Conference on Computer
  Vision and Pattern Recognition. pp. 764--773 (2017)

\bibitem{towardsfairerdatasets}
Yang, K., Qinami, K., Fei-Fei, L., Deng, J., Russakovsky, O.: Towards fairer
  datasets: Filtering and balancing the distribution of the people subtree in
  the imagenet hierarchy. In: Proceedings of the 2020 Conference on Fairness,
  Accountability, and Transparency. pp. 547--558 (2020)

\bibitem{yang2019my}
Yang, X., Mei, H., Xu, K., Wei, X., Yin, B., Lau, R.W.: Where is my mirror? In:
  Proceedings of the IEEE/CVF International Conference on Computer Vision. pp.
  8809--8818 (2019)

\bibitem{yu2015multi}
Yu, F., Koltun, V.: Multi-scale context aggregation by dilated convolutions.
  In: International Conference on Learning Representations (2016)

\bibitem{zhao2017fully}
Zhao, C., Sun, L., Stolkin, R.: A fully end-to-end deep learning approach for
  real-time simultaneous 3{D} reconstruction and material recognition. In: 2017
  18th International Conference on Advanced Robotics (ICAR). pp. 75--82. IEEE
  (2017)

\bibitem{zhao2017pyramid}
Zhao, H., Shi, J., Qi, X., Wang, X., Jia, J.: Pyramid scene parsing network.
  In: Proceedings of the IEEE conference on computer vision and pattern
  recognition. pp. 2881--2890 (2017)

\bibitem{places365}
Zhou, B., Lapedriza, A., Khosla, A., Oliva, A., Torralba, A.: Places: A 10
  million image database for scene recognition. IEEE transactions on pattern
  analysis and machine intelligence  \textbf{40}(6),  1452--1464 (2017)

\bibitem{zhou2017scene}
Zhou, B., Zhao, H., Puig, X., Fidler, S., Barriuso, A., Torralba, A.: Scene
  parsing through ade20k dataset. In: Proceedings of the IEEE Conference on
  Computer Vision and Pattern Recognition (2017)

\bibitem{zhou2019semantic}
Zhou, B., Zhao, H., Puig, X., Xiao, T., Fidler, S., Barriuso, A., Torralba, A.:
  Semantic understanding of scenes through the {ADE20K} dataset. International
  Journal of Computer Vision  \textbf{127}(3),  302--321 (2019)

\bibitem{paszke2016enet}
Paszke, A., Chaurasia, A., Kim, S., Culurciello, E.: E{N}et: A deep neural
  network architecture for real-time semantic segmentation. arXiv preprint
  arXiv:1606.02147  (2016)

\end{thebibliography}
\par\vfill\par
\appendix

\section{Dataset Details}

In this section we supplement Section~\ref{sec:dataset} of the main paper.

In Table~\ref{tab:matarea} we list names used in annotation tools. For brevity, names in the main paper are shortened and ``Photograph/painting'' is called \matlabel{artwork}. We also report the number of images in which a material occurs and total area, the sum over all images of the fraction of pixels covered by a material.

In Table~\ref{tab:fusedpixelcount} we show the number of annotated pixels for each class. This count is according to the resized images which are smaller than the original images.

{
\begin{longtable}{@{}lrrrrcrrrr@{}}
\caption{{\bf Material occurrence.} We report the number of images and total area (in units of image proportion, rounded).}
\label{tab:matarea}\\
\toprule
\endfirsthead
\caption{continued from previous page}\\
\endhead
                               & \multicolumn{4}{c}{Image Count} & \phantom{\;} & \multicolumn{4}{c}{Total Area}\\
\cmidrule{2-5} \cmidrule{7-10}
                               &  All    &   Train &     Val &  Test & &    All  &   Train &     Val &    Test\\\midrule
Animal skin                    &   1,007 &     479 &     260 &     268 & &      34 &      14 &       8 &      11\\
Bone/teeth/horn                &   3,751 &   2,084 &     858 &     809 & &       4 &       2 &       1 &       2\\
Brickwork                      &   1,654 &     862 &     388 &     404 & &     204 &     113 &      46 &      44\\
Cardboard                      &   3,150 &   1,773 &     681 &     696 & &     133 &      73 &      30 &      30\\
Carpet/rug                     &   9,516 &   5,470 &   2,073 &   1,973 & &     985 &     567 &     208 &     209\\
Ceiling tile                   &   2,524 &   1,460 &     529 &     535 & &     299 &     173 &      65 &      61\\
Ceramic                        &   8,314 &   4,608 &   1,854 &   1,852 & &     260 &     135 &      69 &      56\\
Chalkboard/blackboard\;\;      &     668 &     332 &     166 &     170 & &      68 &      34 &      16 &      19\\
Clutter                        &     128 &      41 &      43 &      44 & &      12 &       3 &       5 &       5\\
Concrete                       &   2,853 &   1,381 &     731 &     741 & &     400 &     186 &     109 &     105\\
Cork/corkboard                 &     273 &     122 &      78 &      73 & &       9 &       4 &       2 &       3\\
Engineered stone               &     299 &     134 &      81 &      84 & &      18 &       8 &       5 &       5\\
Fabric/cloth                   &  31,489 &  17,727 &   6,875 &   6,887 & &   4,799 &   2,732 &   1,038 &   1,030\\
Fiberglass wool                &      33 &      12 &       9 &      12 & &       3 &       1 &       1 &       1\\
Fire                           &     412 &     184 &     110 &     118 & &      12 &       5 &       4 &       3\\
Foliage                        &  11,384 &   5,902 &   2,714 &   2,768 & &   1,377 &     640 &     372 &     364\\
Food                           &   2,908 &   1,553 &     687 &     668 & &     287 &     126 &      82 &      79\\
Fur                            &   1,567 &     761 &     398 &     408 & &     206 &      95 &      55 &      55\\
Gemstone/quartz                &     369 &     165 &      99 &     105 & &      10 &       5 &       2 &       3\\
Glass                          &  28,934 &  16,142 &   6,378 &   6,414 & &   2,159 &   1,192 &     488 &     479\\
Hair                           &  17,766 &  10,076 &   3,823 &   3,867 & &     336 &     190 &      74 &      72\\
Ice                            &      96 &      31 &      32 &      33 & &      27 &      10 &       8 &       8\\
Leather                        &   7,354 &   4,146 &   1,609 &   1,599 & &     210 &     118 &      50 &      42\\
Liquid, non-water              &     294 &     129 &      83 &      82 & &       9 &       2 &       4 &       3\\
Metal                          &  30,504 &  16,917 &   6,801 &   6,786 & &     805 &     427 &     187 &     190\\
Mirror                         &   3,242 &   1,871 &     684 &     687 & &     315 &     176 &      67 &      72\\
Paint/plaster/enamel           &  39,323 &  21,765 &   8,773 &   8,785 & &  10,965 &   6,073 &   2,434 &   2,458\\
Paper                          &  20,763 &  11,692 &   4,592 &   4,479 & &     883 &     485 &     200 &     199\\
Pearl                          &     282 &     129 &      77 &      76 & &       0 &       0 &       0 &       0\\
Photograph/painting            &   4,344 &   2,435 &     976 &     933 & &     174 &      90 &      41 &      43\\
Plastic, clear                 &   6,431 &   3,583 &   1,425 &   1,423 & &     129 &      69 &      28 &      31\\
Plastic, non-clear             &  30,506 &  17,154 &   6,662 &   6,690 & &   1,278 &     708 &     282 &     288\\
Rubber/latex                   &   7,811 &   4,244 &   1,788 &   1,779 & &      65 &      32 &      17 &      16\\
Sand                           &     272 &     110 &      76 &      86 & &      70 &      24 &      20 &      26\\
Skin/lips                      &  18,524 &  10,444 &   4,014 &   4,066 & &     509 &     287 &     113 &     108\\
Sky                            &   3,306 &   1,447 &     911 &     948 & &   1,020 &     435 &     286 &     298\\
Snow                           &     191 &      70 &      60 &      61 & &      57 &      19 &      20 &      18\\
Soap                           &     154 &      58 &      50 &      46 & &       0 &       0 &       0 &       0\\
Soil/mud                       &   1,855 &     860 &     495 &     500 & &     165 &      73 &      42 &      51\\
Sponge                         &     326 &     149 &      89 &      88 & &       1 &       1 &       0 &       0\\
Stone, natural                 &   2,076 &     962 &     569 &     545 & &     355 &     156 &     102 &      98\\
Stone, polished                &   1,831 &     993 &     435 &     403 & &     187 &      97 &      46 &      44\\
Styrofoam                      &      88 &      33 &      27 &      28 & &       2 &       1 &       0 &       1\\
Tile                           &  10,173 &   5,722 &   2,206 &   2,245 & &   1,490 &     845 &     321 &     323\\
Wallpaper                      &   1,076 &     577 &     252 &     247 & &     233 &     127 &      56 &      49\\
Water                          &   2,063 &     959 &     552 &     552 & &     564 &     260 &     156 &     149\\
Wax                            &   1,107 &     578 &     260 &     269 & &       7 &       3 &       2 &       2\\
Whiteboard                     &   1,171 &     642 &     265 &     264 & &     111 &      60 &      24 &      27\\
Wicker                         &   1,895 &   1,031 &     438 &     426 & &      75 &      35 &      22 &      18\\
Wood                           &  24,248 &  13,496 &   5,309 &   5,443 & &   3,608 &   2,006 &     802 &     800\\
Wood, tree                     &   2,026 &     929 &     561 &     536 & &      72 &      30 &      19 &      22\\
Asphalt                        &     474 &     211 &     132 &     131 & &      73 &      35 &      17 &      22\\
\bottomrule
\end{longtable}
}

\begin{table}[t]
\centering
\caption{{\bf Material occurrence in pixels.} We report the number of pixels covered by each label according to the resized images used by annotation tools.}
\label{tab:fusedpixelcount}
\begin{tabular}{@{}lrp{3mm}lr@{}}\toprule
Animal skin                    &    22,995,883  & & Paint/plaster/enamel  &  7,796,144,397\\
Bone/teeth/horn                &     3,050,548  & & Paper                 &   628,009,751\\
Brickwork                      &   145,410,237  & & Pearl                 &      411,455\\
Cardboard                      &    93,881,191  & & Photograph/painting   &   123,296,052\\
Carpet/rug                     &   707,147,207  & & Plastic, clear        &    93,002,805\\
Ceiling tile                   &   216,289,692  & & Plastic, non-clear    &   906,618,216\\
Ceramic                        &   185,191,692  & & Rubber/latex          &    45,644,757\\
Chalkboard/blackboard\;\;      &    48,346,203  & & Sand                  &    47,860,125\\
Clutter                        &     8,845,550  & & Skin/lips             &   359,727,474\\
Concrete                       &   283,303,562  & & Sky                   &   702,864,398\\
Cork/corkboard                 &     6,468,131  & & Snow                  &    40,936,881\\
Engineered stone               &    13,140,139  & & Soap                  &      265,782\\
Fabric/cloth                   &  3,408,488,743  & & Soil/mud             &   114,322,155\\
Fiberglass wool                &     1,874,005  & & Sponge                &     1,075,671\\
Fire                           &     7,965,989  & & Stone, natural        &   253,271,347\\
Foliage                        &   961,103,715  & & Stone, polished       &   134,425,626\\
Food                           &   192,755,372  & & Styrofoam             &     1,552,343\\
Fur                            &   145,359,760  & & Tile                  &  1,068,909,615\\
Gemstone/quartz                &     7,273,649  & & Wallpaper             &   168,289,772\\
Glass                          &  1,535,538,311  & & Water                &   390,040,955\\
Hair                           &   238,600,730  & & Wax                   &     4,791,692\\
Ice                            &    18,308,742  & & Whiteboard            &    80,692,711\\
Leather                        &   149,122,712  & & Wicker                &    50,066,493\\
Liquid, non-water              &     5,861,652  & & Wood                  &  2,584,799,129\\
Metal                          &   573,827,793  & & Wood, tree            &    50,922,547\\
Mirror                         &   224,631,105  & & Asphalt               &    51,218,822\\
\bottomrule
\end{tabular}
\end{table}

\begin{table}[t]
\centering
\caption{{\bf Case resolution.} For some cases we provided additional instruction, which we summarize here.}
\label{tab:labeldesc}
\begin{tabular}{@{}lrl@{}}\toprule
Case                      & & Resolution\\\midrule
Skin with sparse hair     & & \matlabel{Skin} for people; \matlabel{animal skin} for animals.\\
Coat of hair (\eg, horse) & & \matlabel{Fur}.\\
Smoothed stone            & & \matlabel{Polished stone}.\\
Laminated paper           & & \matlabel{Clear plastic}.\\
Sauces                    & & \matlabel{Food} on food; \matlabel{non-water liquid} during preparation.\\
Chandelier prisms         & & \matlabel{Gemstone} or \matlabel{glass} based on appearance.\\
Seasoned or blued metal   & & \matlabel{Metal}.\\
Metal patina              & & \matlabel{Metal}.\\
Printed text              & & The underlying material.\\
Mirror-like finishes      & & \matlabel{Mirror} if sole purpose is to reflect; the material otherwise.\\
Wrapped items             & & The material of the wrap.\\
Electronic display        & & \matlabel{Glass}.\\
Glass-top surface         & & \matlabel{Glass}.\\
Thatch                    & & \matlabel{Wicker}.\\
Stained wood              & & \matlabel{Wood}.\\
Projection screen         & & \matlabel{Not on list}.\\
Vinyl                     & & The closest of \matlabel{non-clear plastic}, \matlabel{rubber} or \matlabel{leather}.\\
\bottomrule
\end{tabular}
\end{table}

\begin{table}[t]
\centering
\caption{{\bf Objects and functional spaces.} We report the number of images for the largest classes of detected objects ({\it top}) and estimated scene functions ({\it bottom}).}
\label{tab:topk}
\begin{tabular}{@{}lrrrrp{3mm}lrrrr@{}}\toprule
                     &  All   &  Train &    Val & Test   & &                      &   All  &  Train &    Val &   Test\\\midrule
person               & 19,966 & 11,219 &  4,303 &  4,426 & & tie                  &  1,398 &    802 &    280 &    314\\
chair                & 17,617 &  9,987 &  3,826 &  3,780 & & bench                &  1,196 &    671 &    244 &    277\\
dining table         &  8,086 &  4,511 &  1,765 &  1,806 & & keyboard             &  1,192 &    648 &    272 &    272\\
bottle               &  5,964 &  3,320 &  1,313 &  1,325 & & cell phone           &  1,121 &    629 &    269 &    222\\
cup                  &  5,656 &  3,136 &  1,248 &  1,265 & & mouse                &    939 &    516 &    199 &    224\\
potted plant         &  5,078 &  2,762 &  1,122 &  1,191 & & refrigerator         &    834 &    504 &    161 &    168\\
book                 &  4,384 &  2,465 &    976 &    939 & & backpack             &    739 &    420 &    154 &    165\\
tv                   &  4,303 &  2,411 &    947 &    942 & & oven                 &    737 &    399 &    173 &    165\\
laptop               &  3,076 &  1,737 &    664 &    675 & & remote               &    718 &    403 &    166 &    148\\
bowl                 &  2,900 &  1,579 &    636 &    682 & & dog                  &    692 &    369 &    162 &    160\\
couch                &  2,846 &  1,614 &    628 &    602 & & cat                  &    685 &    344 &    162 &    178\\
vase                 &  2,790 &  1,551 &    626 &    609 & & toilet               &    677 &    383 &    144 &    149\\
bed                  &  2,357 &  1,348 &    524 &    482 & & knife                &    579 &    335 &    123 &    120\\
sink                 &  1,747 &    949 &    395 &    402 & & car                  &    542 &    292 &    128 &    121\\
handbag              &  1,617 &    906 &    366 &    345 & & boat                 &    524 &    227 &    136 &    161\\
wine glass           &  1,473 &    797 &    332 &    343 & & suitcase             &    510 &    310 &     94 &    106\\
clock                &  1,452 &    814 &    294 &    343 & & spoon                &    477 &    258 &    106 &    112\\\midrule
working              & 14,343 &  8,032 &  3,124 &  3,166 & & swimming             &    868 &    397 &    240 &    230\\
reading              & 14,039 &  7,931 &  3,118 &  2,970 & & sports               &    824 &    442 &    181 &    198\\
socializing          &  8,545 &  4,869 &  1,794 &  1,873 & & using tools          &    686 &    369 &    149 &    167\\
congregating         &  7,317 &  4,129 &  1,559 &  1,620 & & praying              &    649 &    363 &    144 &    138\\
eating               &  5,862 &  3,217 &  1,294 &  1,345 & & touring              &    626 &    283 &    159 &    180\\
shopping             &  2,419 &  1,325 &    563 &    526 & & waiting in line      &    593 &    362 &    118 &    113\\
studying             &  2,070 &  1,147 &    459 &    463 & & exercise             &    574 &    329 &    106 &    137\\
competing            &  1,960 &  1,085 &    410 &    458 & & diving               &    556 &    275 &    163 &    117\\
spectating           &  1,489 &    845 &    305 &    335 & & bathing              &    524 &    288 &    120 &    115\\
training             &  1,335 &    744 &    295 &    295 & & research             &    451 &    251 &     92 &    108\\
transporting         &  1,153 &    587 &    268 &    297 & & cleaning             &    445 &    247 &     94 &    104\\
boating              &    876 &    371 &    235 &    267 & & driving              &    404 &    199 &     92 &    113\\
\bottomrule
\end{tabular}
\end{table}

\begin{table}[t]
\centering
\caption{{\bf Judgments.} We report the number of unique opinions (\ie, label maps) collected for images.}
\label{tab:judgments}
\begin{tabular}{@{}ccr@{}}\toprule
Label Map Count   & &  Images\\\midrule
1                 & &  1,245\\
2                 & & 35,039\\
3                 & &  7,459\\
4                 & &    122\\
5                 & &    867\\
\bottomrule
\end{tabular}
\end{table}

\begin{figure}[t]
\includegraphics[height=13.5ex]{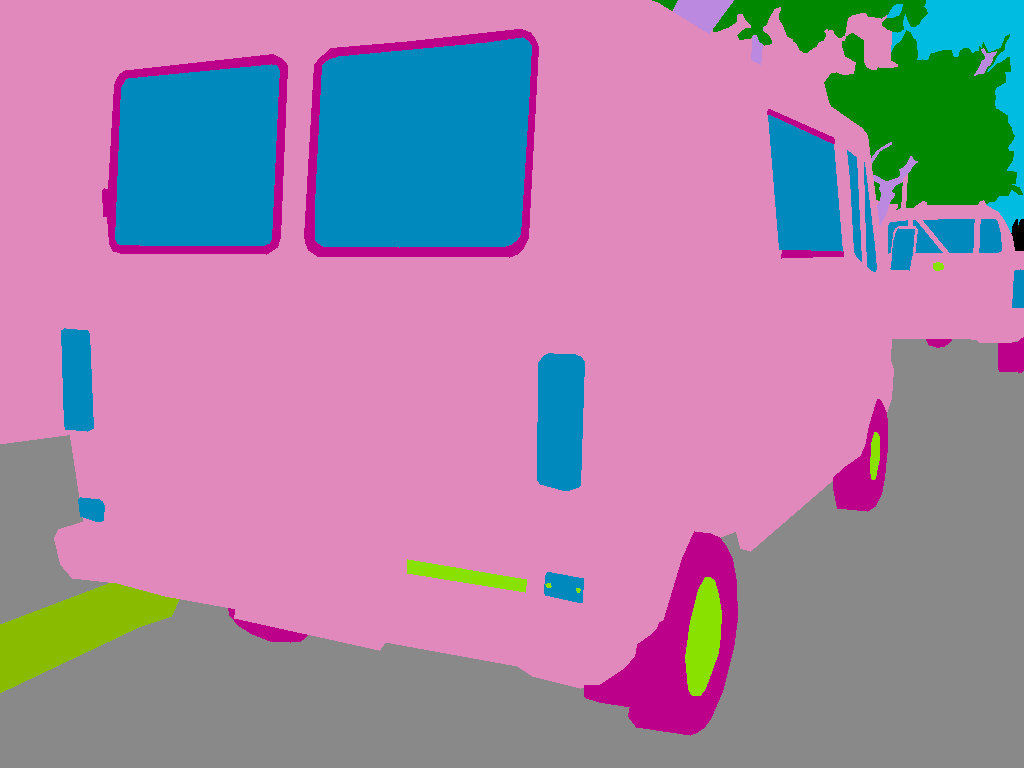}\hfill
\includegraphics[height=13.5ex]{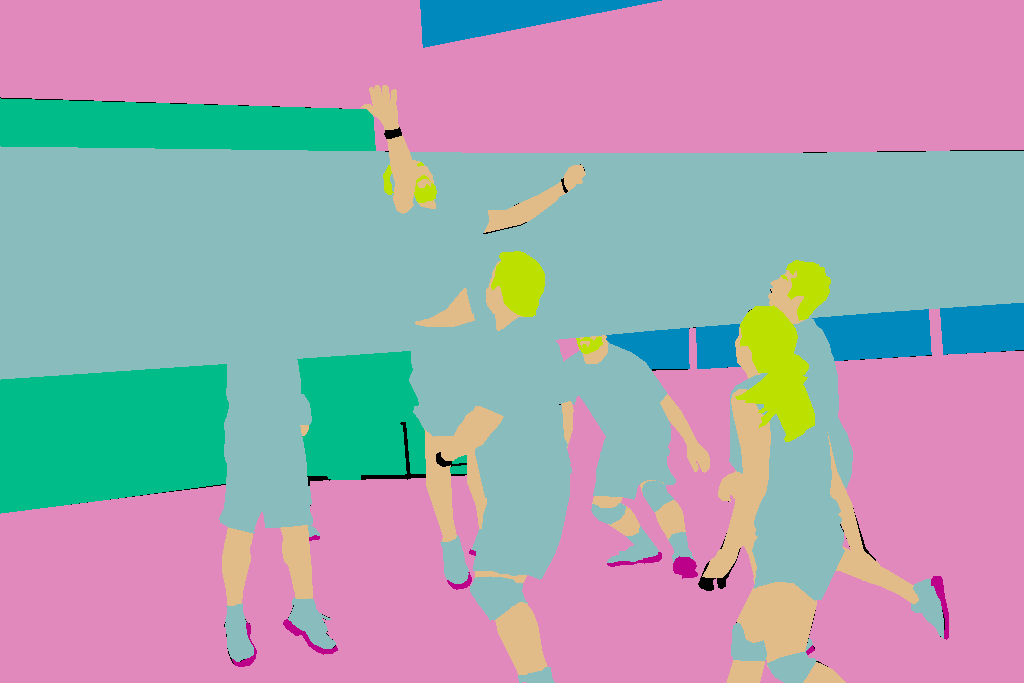}\hfill
\includegraphics[height=13.5ex]{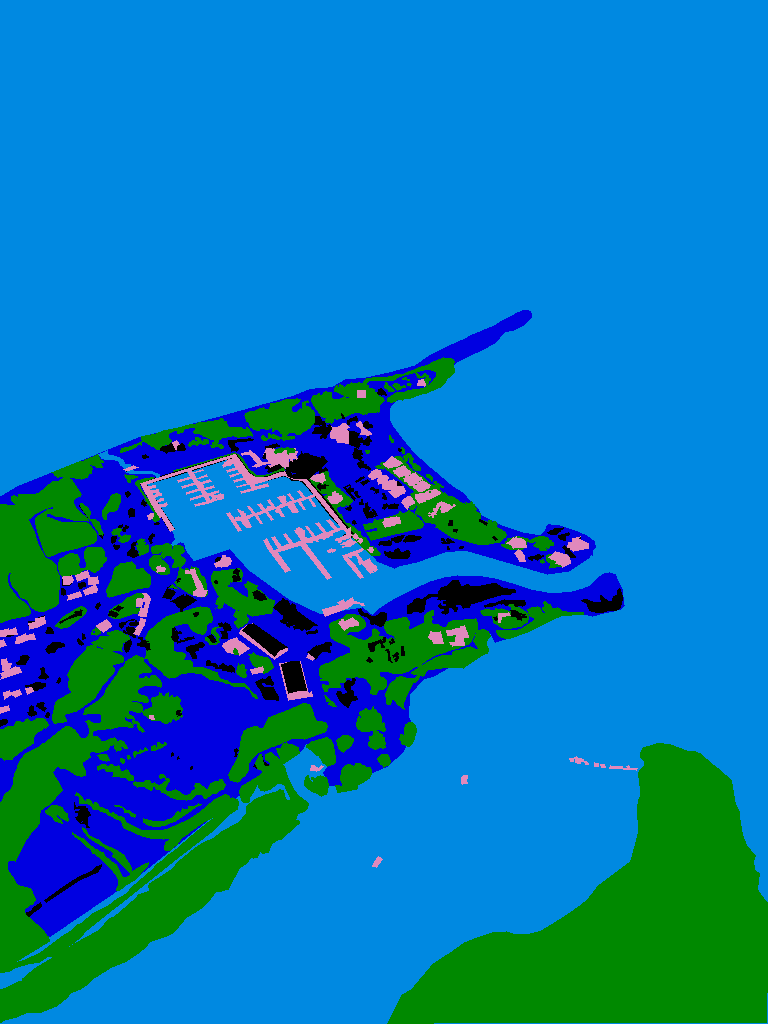}\hfill
\includegraphics[height=13.5ex]{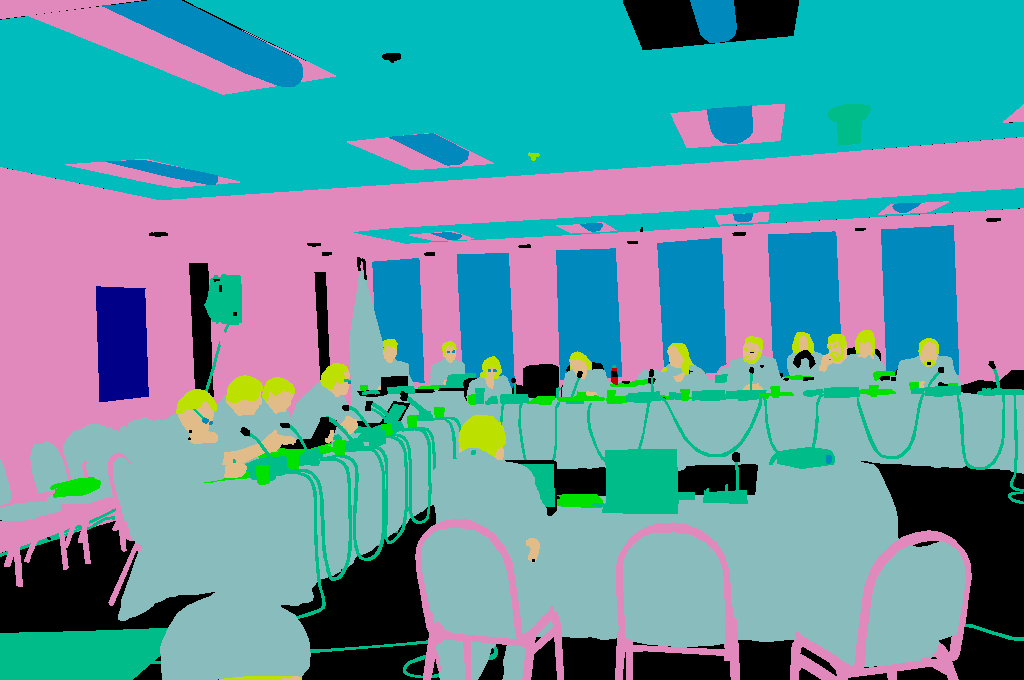}\hfill
\includegraphics[height=13.5ex]{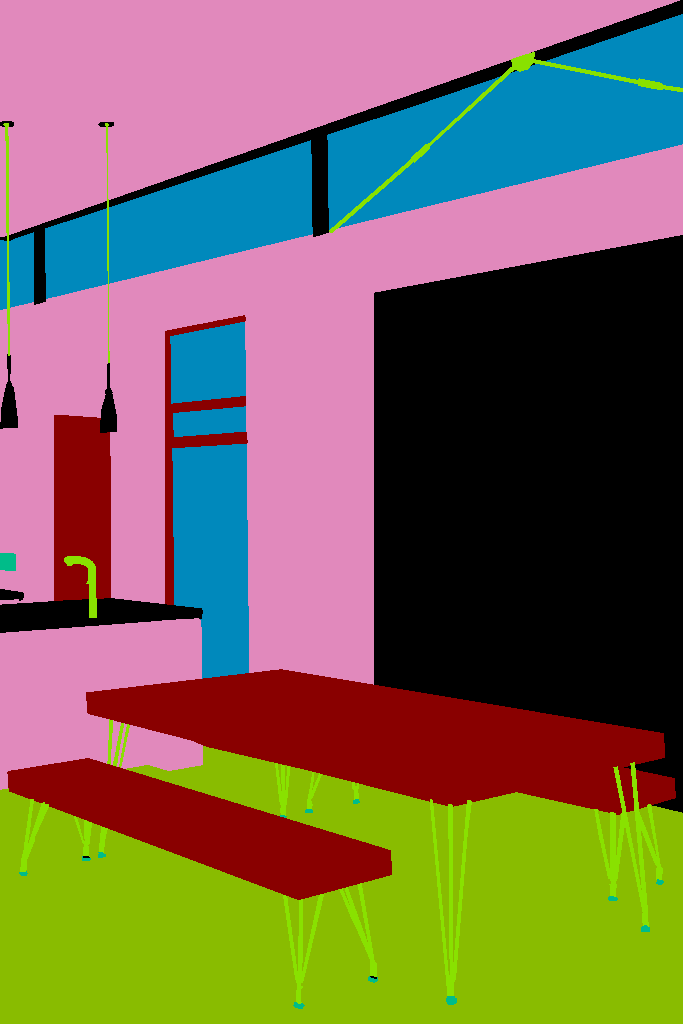}\hfill
\crule{0.15ex}{13.5ex}\hfill
\includegraphics[height=13.5ex]{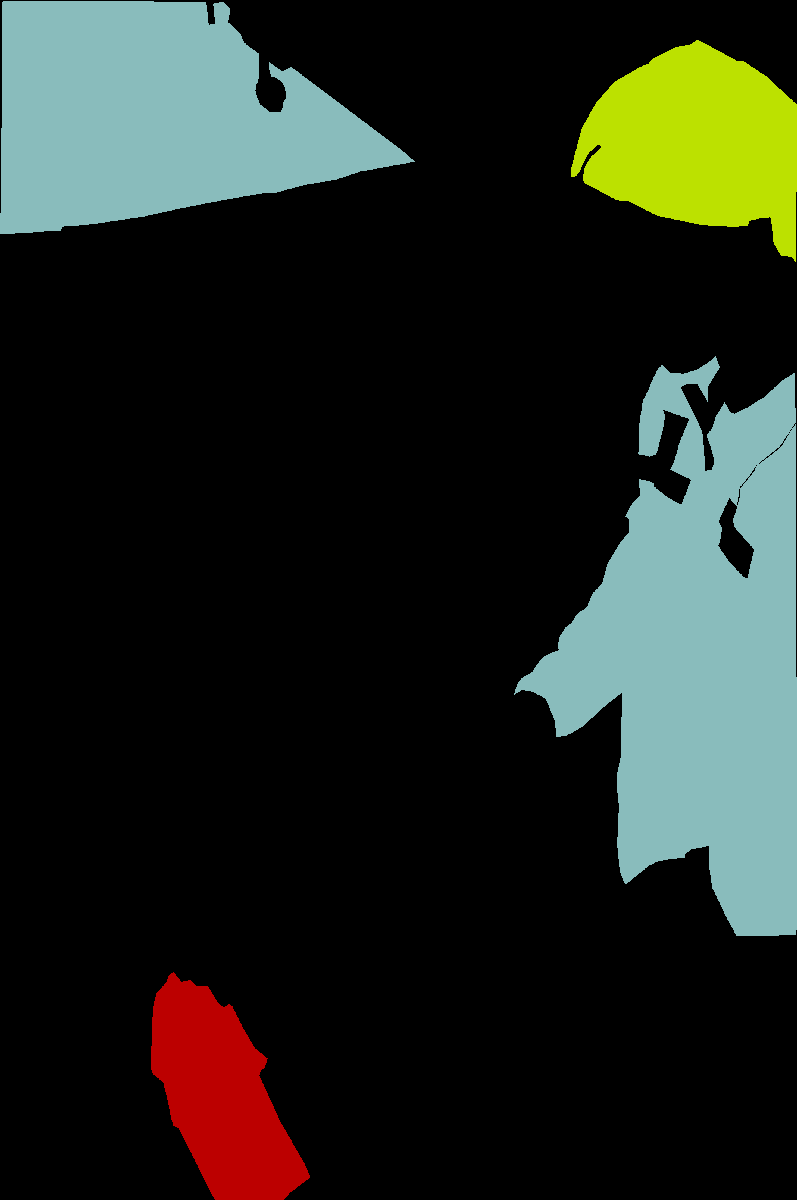}
\caption{{\bf Fused material labels.} {\it Left to right:} van, sports, aerial photo, conference and dining area. The 5th image has a label density close to the mean density of \OM. The rightmost image is a fused label map from OpenSurfaces with a label density close to the mean density of OpenSurfaces. See Table~\ref{tab:fusedcount} for color legend.}
\label{fig:morefusedlabels}
\end{figure}

We found that asking annotators to label all surfaces required extensive instruction. Our training document grew to include clarifications for rare and uncommon cases. In Table~\ref{tab:labeldesc} we summarize how we choose to resolve cases.

In Table~\ref{tab:topk} we report the number of images in which an object class is detected by~\cite{detectron}, and the number of images which are predicted by~\cite{places365} to have scene elements for an activity. There are 80 object classes and 30 functional scene attributes. For brevity, we report only the largest classes.

For most images we collected two unique opinions for labels. In Table~\ref{tab:judgments} we report the number of images with a given number of opinions.

In Figure~\ref{fig:morefusedlabels} we expand on Figure~\ref{fig:fusedlabels} by showing more fused label maps and we show a fused label map from \OM and OpenSurfaces which are representative of the mean density of the respective datasets.

\section{Skin Type Experiment}
In Section~\ref{sec:skin}, we compared skin accuracies for three skin groups, Type I-II, Type III-IV, and Type V-VI. In order to compute accuracy we have to assign ground truth pixels to a group. We do this for images which contain detections of only one skin group. However, there are images where multiple skin groups co-occur and where no skin groups were detected. We do not evaluate on these two scenarios to avoid assigning groups incorrectly.

\section{Benchmark Experiment Details}
In this section we include more details on training our material segmentation benchmark model, DMS-46, from Section~\ref{sec:baseline} of the main paper. All the models are trained on NVIDIA Tesla V100 GPUs with 32 GB of memory.

\subsection{Data Augmentation}
In this section we show details on how we apply different data augmentation in training. We apply the following data transformation in order: 

{\bf Scale.} We first scale the input image so that the shortest dimension is 512 given that the training image size has height 512 and width 512. Then we randomly scale the input dimension with a ratio in [1, 2, 3, 4] uniformly.

{\bf Horizontal Flip.} We apply random horizontal flip with probability 0.5. 

{\bf Vertical Flip.} We apply random vertical flip with probability 0.5. 

{\bf Color Jitter.} We apply color jitter with probability 0.9, using torchvision\footnote{https://pytorch.org/vision/} ColorJitter with brightness 0.4, contrast 0.4, saturation 0.4, and hue 0.1. 

{\bf Gaussian Blur or Gaussian Noise.} We apply this transformation with probability 0.5. Gaussian blur or Gaussian noise is selected with equal chance. We use a kernel size of 3 for Gaussian blur with uniformly chosen standard deviation in [0.1, 2.0]. Gaussian noise has mean of 0 and standard deviation 3 across all the pixels.

{\bf Rotation.} We apply random rotation in [-45, 45] degrees with probability 0.5. We fill 0 for the area outside the rotated color image and an ignore value for the rotated segmentation map. The loss calculation ignores those pixels.

{\bf Crop.} Finally, we randomly crop a subregion, height 512 and width 512, to feed into the neural network.

\subsection{Loss Function}
We use weighted symmetric cross entropy~\cite{wang2019symmetric} as the loss function for DMS-46. The weight ${W_{i}}$ for each class is calculated as a function of frequency of pixel count, ${F_{i}}$, for each material class ${i \in N}$~\cite{paszke2016enet}, in Equation~\ref{eq:weighted}.\begin{equation} \label{eq:weighted}
W_{i} = \frac{1} {\log \left(1.02 + \frac{F_{i}}  {\sum_{i=1}^{N}F_{i}}\right)}
\end{equation} 

The number 1.02 is introduced in~\cite{paszke2016enet} to restrict the class weights in [1, 50] as the probability approaches 0. The weights we are using for DMS-46 are presented in Table~\ref{tab:allnames}. 

Symmetric cross entropy (SCE)~\cite{wang2019symmetric} is composed of a regular cross entropy (CE) and a reverse cross entropy (RCE) to avoid overfitting to noisy labels. Given the target distribution P and the predicted distribution Q, Equation~\ref{eq:sce} shows each part of the loss function for SCE. We choose $\alpha=1$ and $\beta=0.5$ for the weighting coefficients.
\begin{equation} \label{eq:sce}
L_{SCE} = \alpha L_{CE} + \beta L_{RCE} = \alpha (- \sum P \log Q \\) + \beta (- \sum Q \log P \\)
\end{equation}

\begin{table}[t]
\centering
\caption{{\bf Class weights.} We show the class weights we applied in the loss function for DMS-46.}
\label{tab:allnames}
\begin{tabular}{@{}lrp{3mm}lrp{3mm}lr@{}}\toprule
Label & Weight                        & & Label & Weight                         & & Label & Weight                          \\\midrule
  
Bone             & 50.259 & & Whiteboard & 43.585 & & Hair & 33.870  \\
Wax              & 50.140 & & Clear plastic & 42.709 & & Water & 30.402  \\
Clutter          & 50.136 & & Soil & 42.585 & & Skin & 29.049  \\
Cork             & 49.995 & & Cardboard & 42.482 & & Sky & 24.133  \\
Fire             & 49.945 & & Artwork & 40.905 & & Metal & 23.981  \\
Gemstone         & 49.826 & & Fur & 40.427 & & Paper & 22.447  \\
Engineered stone & 49.459 & & Pol. stone & 40.226 & & Carpet & 20.422  \\
Ice              & 49.163 & & Brickwork & 38.979 & & Foliage & 19.325  \\
Animal skin      & 48.646 & & Leather & 38.715 & & Non-clear plastic & 17.986  \\
Snow             & 47.972 & & Food & 38.368 & & Tile & 15.895  \\
Sand             & 47.603 & & Wallpaper & 37.854 & & Glass & 12.555  \\
Tree wood        & 46.759 & & Ceramic & 37.201 & & Wood & 8.388  \\
Rubber           & 46.672 & & Nat. stone & 35.919 & & Fabric & 6.596  \\
Wicker           & 46.465 & & Mirror & 34.651 & & Paint & 3.415  \\
Chalkboard       & 46.462 & & Ceiling tile & 34.617  \\
Asphalt          & 46.447 & & Concrete & 34.095  \\

\bottomrule
\end{tabular}
\end{table}

\subsection{Model Architecture Implementation}
We select ResNet50~\cite{he2016deep} with dilated convolutions~\cite{chen2017deeplab,yu2015multi} as the encoder, and Pyramid Pooling Module from PSPNet~\cite{zhao2017pyramid} as the decoder. We choose this architecture because it has been shown to be effective for scene parsing~\cite{zhao2017pyramid,zhou2019semantic}. We use a publicly-available implementation of ResNet50dilated architecture with pre-trained weights (on an ImageNet task) from~\cite{zhou2017scene,zhou2019semantic}\footnote{https://github.com/CSAILVision/semantic-segmentation-pytorch}, under a BSD 3-Clause License.

\subsection{Material Class Selection For Benchmark}
In Section~\ref{sec:baseline} we reported empirically finding that six material categories (\matlabel{non-water liquid}, \matlabel{fiberglass}, \matlabel{sponge}, \matlabel{pearl}, \matlabel{soap} and \matlabel{styrofoam}) fail consistently across models. We present the three top candidates of DMS-52 which led us to this conclusion. Each one is the best fitted model, according to DMS-val, from a comprehensive hyper-parameter search on learning rate, learning rate scheduler, and optimizer.
The first model, called DMS-52, is the best model across all models, is introduced in the main paper, and we report the per-class performance in Table~\ref{tab:first}.
The second model, called DMS-52 variant A, has the same architecture as DMS-52 and uses all of OpenSurfaces data as additional training data. We report the per-class performance of DMS-52A in Table~\ref{tab:second}.
The third model, called DMS-52 variant B, has a ResNet101 architecture and uses OpenSurfaces data as additional training data. We report the per-class performance of DMS-52B in Table~\ref{tab:third}.
Across DMS-52, DMS-52A and DMS-52B the same six material classes are the worst-performing categories. Based on these findings we selected the other 46 categories for a benchmark and leave these six to future work.

\begin{table}[t]
\centering
\caption{{\bf DMS-Val results for DMS-52.} Results are sorted by accuracy.}
\label{tab:first}
\begin{tabular}{@{}lccp{3mm}lccp{3mm}lcc@{}}\toprule
     &   Acc & IoU & &     & Acc & IoU & &   & Acc & IoU\\\midrule
  
Sky & 0.937 & 0.891 & & Glass & 0.703 & 0.489 & & Animal skin & 0.396 & 0.268 \\
Fur & 0.913 & 0.694 & & Paper & 0.686 & 0.496 & & Rubber & 0.345 & 0.240 \\
Foliage & 0.897 & 0.769 & & Leather & 0.676 & 0.397 & & Pol. stone & 0.332 & 0.236 \\
Ceiling tile & 0.890 & 0.679 & & Nat. stone & 0.634 & 0.447 & & Tree wood & 0.327 & 0.224 \\
Hair & 0.885 & 0.673 & & Wax & 0.626 & 0.430 & & Ice & 0.320 & 0.284 \\
Food & 0.882 & 0.689 & & Wicker & 0.622 & 0.432 & & Bone & 0.213 & 0.178 \\
Water & 0.881 & 0.695 & & Wallpaper & 0.603 & 0.397 & & Clutter & 0.209 & 0.186 \\
Skin & 0.876 & 0.647 & & Concrete & 0.579 & 0.333 & & Gemstone & 0.127 & 0.077 \\
Carpet & 0.855 & 0.582 & & Soil & 0.578 & 0.376 & & Cork & 0.115 & 0.102 \\
Fire & 0.821 & 0.621 & & Cardboard & 0.571 & 0.340 & & Eng. stone & 0.096 & 0.069 \\
Wood & 0.801 & 0.657 & & Non-clear plastic & 0.562 & 0.322 & & {\bf Sponge} & 0.051 & 0.050 \\
Fabric & 0.787 & 0.690 & & Asphalt & 0.560 & 0.386 & & {\bf Liquid} & 0.048 & 0.044 \\
Brickwork & 0.785 & 0.514 & & Metal & 0.548 & 0.305 & & {\bf Fiberglass} & 0.034 & 0.034 \\
Whiteboard & 0.771 & 0.508 & & Sand & 0.548 & 0.407 & & {\bf Styrofoam} & 0.003 & 0.003 \\
Tile & 0.752 & 0.564 & & Snow & 0.495 & 0.414 & & {\bf Pearl} & 0.000 & 0.000 \\
Chalkboard & 0.747 & 0.616 & & Clear plastic & 0.441 & 0.254 & & {\bf Soap} & 0.000 & 0.000 \\
Ceramic & 0.746 & 0.482 & & Mirror & 0.423 & 0.297 \\
Paint & 0.707 & 0.640 & & Artwork & 0.407 & 0.271 \\

\bottomrule
\end{tabular}
\end{table}

\begin{table}[t]
\centering
\caption{{\bf DMS-Val results for DMS-52A.} Results are sorted by accuracy.}
\label{tab:second}
\begin{tabular}{@{}lccp{3mm}lccp{3mm}lcc@{}}\toprule
     &   Acc & IoU & &     & Acc & IoU & &   & Acc & IoU\\\midrule
    Sky & 0.946 & 0.889 & & Leather & 0.695 & 0.407 & & Clear plastic & 0.405 & 0.255 \\
Fur & 0.921 & 0.692 & & Paint & 0.680 & 0.625 & & Rubber & 0.367 & 0.240 \\
Foliage & 0.912 & 0.768 & & Wicker & 0.670 & 0.436 & & Tree wood & 0.358 & 0.221 \\
Ceiling tile & 0.886 & 0.686 & & Concrete & 0.646 & 0.347 & & Wax & 0.327 & 0.246 \\
Hair & 0.883 & 0.677 & & Soil & 0.635 & 0.385 & & Ice & 0.230 & 0.228 \\
Water & 0.883 & 0.707 & & Fire & 0.626 & 0.570 & & Eng. stone & 0.207 & 0.108 \\
Skin & 0.877 & 0.636 & & Nat. stone & 0.620 & 0.439 & & Clutter & 0.204 & 0.185 \\
Food & 0.875 & 0.688 & & Wallpaper & 0.600 & 0.417 & & Bone & 0.167 & 0.139 \\
Carpet & 0.830 & 0.614 & & Asphalt & 0.599 & 0.401 & & Cork & 0.126 & 0.112 \\
Wood & 0.821 & 0.654 & & Cardboard & 0.586 & 0.362 & & Gemstone & 0.087 & 0.057 \\
Fabric & 0.801 & 0.700 & & Snow & 0.584 & 0.484 & & {\bf Sponge} & 0.066 & 0.060 \\
Whiteboard & 0.801 & 0.515 & & Non-clear plastic & 0.555 & 0.319 & & {\bf Fiberglass} & 0.029 & 0.029 \\
Brickwork & 0.789 & 0.496 & & Metal & 0.548 & 0.289 & & {\bf Liquid} & 0.009 & 0.009 \\
Ceramic & 0.772 & 0.471 & & Animal skin & 0.517 & 0.272 & & {\bf Pearl} & 0.000 & 0.000 \\
Tile & 0.745 & 0.576 & & Pol. stone & 0.489 & 0.254 & & {\bf Soap} & 0.000 & 0.000 \\
Chalkboard & 0.744 & 0.593 & & Sand & 0.463 & 0.389 & & {\bf Styrofoam} & 0.000 & 0.000 \\
Paper & 0.718 & 0.509 & & Artwork & 0.445 & 0.294 \\
Glass & 0.696 & 0.502 & & Mirror & 0.434 & 0.308 \\
\bottomrule
\end{tabular}
\end{table}

\begin{table}[t]
\centering
\caption{{\bf DMS-Val results for DMS-52B.} Results are sorted by accuracy.}
\label{tab:third}
\begin{tabular}{@{}lccp{3mm}lccp{3mm}lcc@{}}\toprule
     &   Acc & IoU & &     & Acc & IoU & &   & Acc & IoU\\\midrule
Sky & 0.943 & 0.865 & & Glass & 0.690 & 0.488 & & Tree wood & 0.352 & 0.257 \\
Foliage & 0.905 & 0.776 & & Nat. stone & 0.685 & 0.402 & & Rubber & 0.310 & 0.265 \\
Hair & 0.891 & 0.687 & & Wicker & 0.684 & 0.454 & & Animal skin & 0.301 & 0.254 \\
Water & 0.889 & 0.655 & & Paper & 0.681 & 0.510 & & Ice & 0.239 & 0.232 \\
Food & 0.862 & 0.687 & & Wallpaper & 0.651 & 0.384 & & Bone & 0.206 & 0.177 \\
Skin & 0.861 & 0.675 & & Leather & 0.603 & 0.431 & & Wax & 0.202 & 0.166 \\
Ceiling tile & 0.858 & 0.673 & & Snow & 0.593 & 0.507 & & Eng. stone & 0.198 & 0.106 \\
Carpet & 0.847 & 0.566 & & Concrete & 0.587 & 0.316 & & Cork & 0.192 & 0.134 \\
Fur & 0.829 & 0.720 & & Metal & 0.553 & 0.300 & & Clutter & 0.131 & 0.113 \\
Wood & 0.820 & 0.642 & & Soil & 0.542 & 0.337 & & Gemstone & 0.095 & 0.082 \\
Fabric & 0.789 & 0.701 & & Non-clear plastic & 0.540 & 0.344 & & {\bf Liquid} & 0.029 & 0.022 \\
Whiteboard & 0.752 & 0.539 & & Asphalt & 0.536 & 0.369 & & {\bf Fiberglass} & 0.017 & 0.016 \\
Fire & 0.739 & 0.654 & & Cardboard & 0.529 & 0.367 & & {\bf Sponge} & 0.003 & 0.003 \\
Ceramic & 0.737 & 0.499 & & Sand & 0.498 & 0.407 & & {\bf Pearl} & 0.000 & 0.000 \\
Brickwork & 0.734 & 0.501 & & Pol. stone & 0.459 & 0.238 & & {\bf Soap} & 0.000 & 0.000 \\
Chalkboard & 0.733 & 0.634 & & Artwork & 0.438 & 0.276 & & {\bf Styrofoam} & 0.000 & 0.000 \\
Paint & 0.705 & 0.633 & & Clear plastic & 0.392 & 0.251 \\
Tile & 0.704 & 0.535 & & Mirror & 0.358 & 0.265 \\

\bottomrule
\end{tabular}
\end{table}

\subsection{More Real-World Examples}
\begin{figure}[t]
\includegraphics[height=14.7ex]{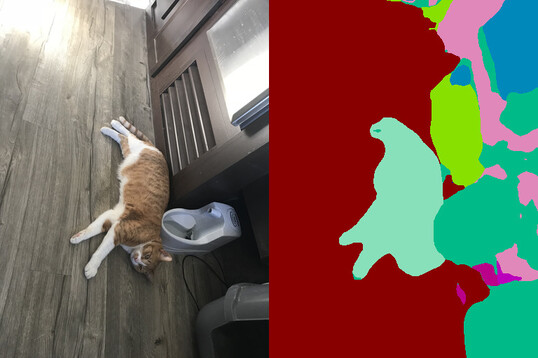}\hfill
\includegraphics[height=14.7ex]{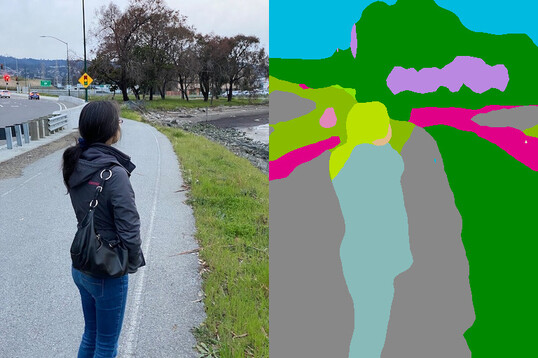}\hfill
\includegraphics[height=14.7ex]{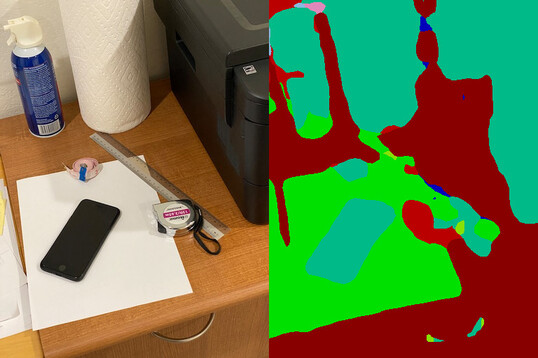}\hfill
\includegraphics[height=14.7ex]{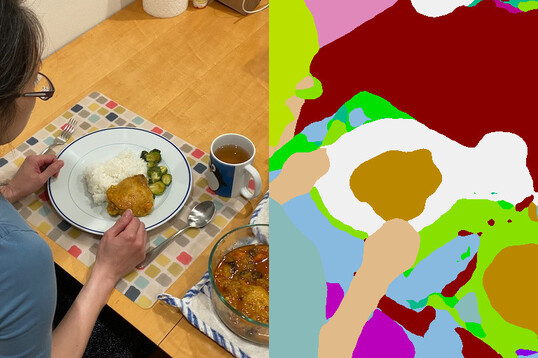}
\includegraphics[height=13.0ex]{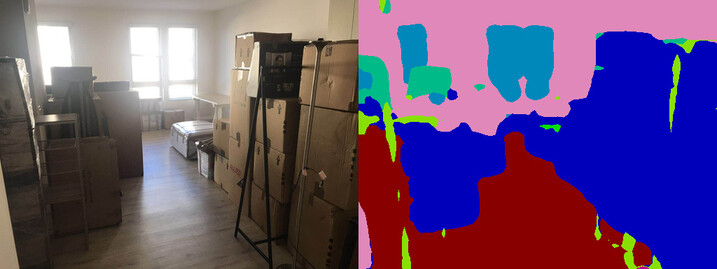}\hfill
\includegraphics[height=13.0ex]{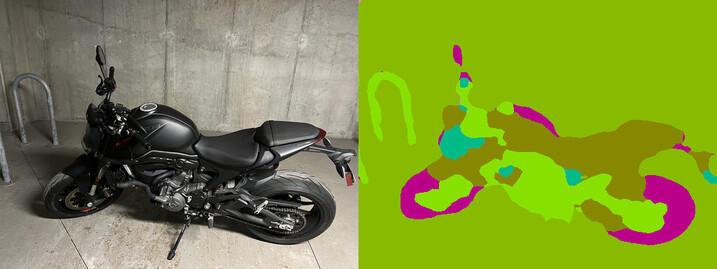}\hfill
\includegraphics[height=13.0ex]{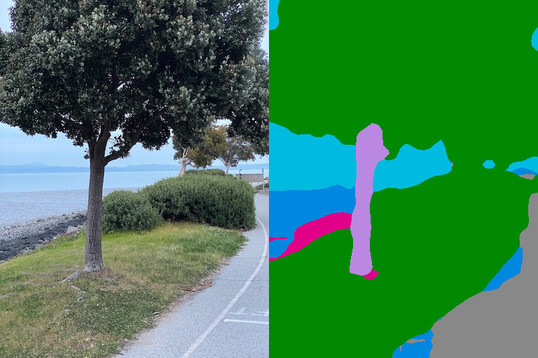} 
\includegraphics[height=13.55ex]{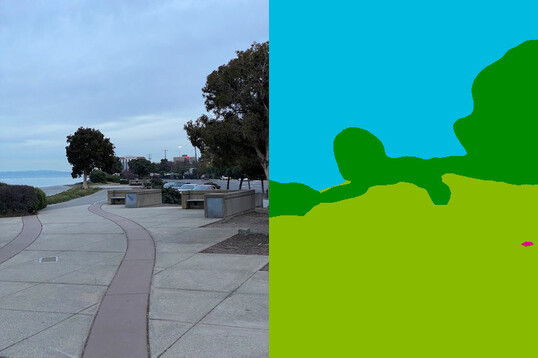}\hfill
\includegraphics[height=13.55ex]{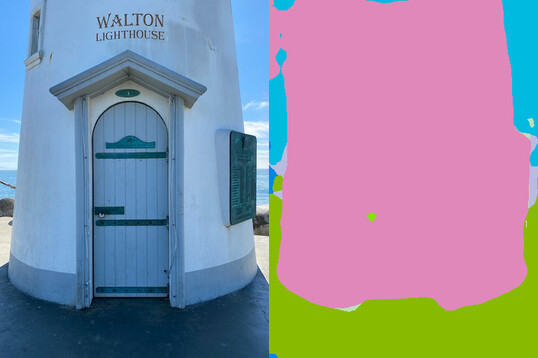}\hfill
\includegraphics[height=13.55ex]{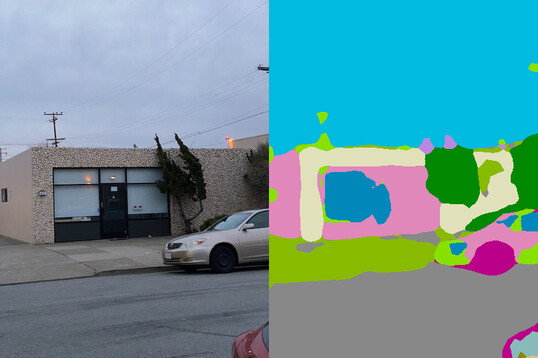}\hfill
\includegraphics[height=13.55ex]{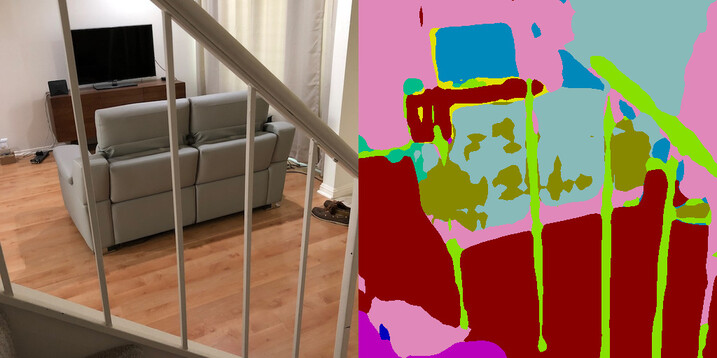}
\caption{{\bf Real-world examples.} Our model, \OM-46, predicts 46 kinds of indoor and outdoor materials. See Table~\ref{tab:fusedcount} for color legend.}
\label{fig:examples2}
\end{figure}

We show more DMS-46 predictions on real world images in Figure~\ref{fig:examples2}.

\section{Image Credits}

Photos in the paper and supplemental are used with permission. We thank the following Flickr users for sharing their photos with a CC-BY-2.0\footnote{https://creativecommons.org/licenses/by/2.0/} license. Some photos in the main paper were changed to remove logos or faces, scale, mask, or crop.

Image credits: 
Random Retail,
Ross Harmes,
Amazing Almonds,
Jonathan Hetzel,
Patrick Lentz,
Colleen Benelli,
Jannes Pockele,
FaceMePLS,
Michael Button,
samuelrodgers752,
Ron Cogswell,
David Costa,
Janet McKnight,
Jennifer,
Adam Bartlett,
www.toprq.com/iphone,
Seth Goodman,
Municipalidad Antofagasta,
Tom Hughes-Croucher,
Travis Grathwell,
Associated Fabrication,
Tjeerd Wiersma,
mike.benedetti,
Frédéric BISSON,
Wendy Cutler,
with wind,
Barry Badcock,
Joel Kramer,
Gwydion M. Williams,
Andreas Kontokanis,
Jim Winstead,
Mike Mozart,
Keith Cooper,
Kurman Communications, Inc.,
Paragon Apartments,
Pedro Ribeiro Simões,
jojo nicdao,
Gobierno Cholula,
David Becker,
Emmanuel DYAN,
Ewen Roberts,
Supermac1961,
fugzu,
Erik (HASH) Hersman,
Eugene Kim,
Bernt Rostad,
andrechinn,
Geología Valdivia,
peapod labs,
Alex Indigo,
Turol Jones, un artista de cojones,
Blake Patterson,
cavenderamy,
tapetenpics,
DLSimaging,
Andy / Andrew Fogg,
Scott,
Justin Ruckman,
espring4224,
objectivised,
Li-Ji,
Bruno Kussler Marques,
and BurnAway.

\end{document}